\pdfoutput=1

\documentclass[11pt]{article}

\usepackage[final]{coling}

\usepackage{times}
\usepackage{latexsym}
\usepackage{graphicx}
\usepackage{url}

\usepackage[T1]{fontenc}

\usepackage[utf8]{inputenc}
\usepackage{multirow}
\usepackage{booktabs} 

\usepackage{microtype}
\usepackage{float}

\usepackage{inconsolata}
\usepackage{booktabs}
\usepackage{subcaption}

\usepackage{amssymb}
\usepackage{stfloats}
\usepackage{multicol}
\usepackage{tabularx} 
\title{Instruction Tuning Vs. In-Context Learning: Revisiting Large Language Models in Few-Shot Computational Social Science}

\author{Taihang Wang\textsuperscript{$\clubsuit$}, Xiaoman Xu\textsuperscript{$\clubsuit$}, Yimin Wang\textsuperscript{$\heartsuit$} \and Ye Jiang\textsuperscript{$\clubsuit$}\thanks{Corresponding author} \\
        College of Information Science and Technology\textsuperscript{$\clubsuit$} \\ 
        College of Data Science\textsuperscript{$\heartsuit$} \\
        Qingdao University of Science and Technology \\ China}

\begin{document}
\maketitle
\begin{abstract}
Real-world applications of large language models (LLMs) in computational social science (CSS) tasks primarily depend on the effectiveness of instruction tuning (IT) or in-context learning (ICL). While IT has shown highly effective at fine-tuning LLMs for various tasks, ICL offers a rapid alternative for task adaptation by learning from examples without explicit gradient updates. In this paper, we evaluate the classification performance of LLMs using IT versus ICL in few-shot CSS tasks. The experimental results indicate that ICL consistently outperforms IT in most CSS tasks. Additionally, we investigate the relationship between the increasing number of training samples and LLM performance. Our findings show that simply increasing the number of samples without considering their quality does not consistently enhance the performance of LLMs with either ICL or IT and can sometimes even result in a performance decline. Finally, we compare three prompting strategies, demonstrating that ICL is more effective than zero-shot and Chain-of-Thought (CoT). Our research highlights the significant advantages of ICL in handling CSS tasks in few-shot settings and emphasizes the importance of optimizing sample quality and prompting strategies to improve LLM classification performance. The code will be made available.


\end{abstract}

\section{Introduction}



Instruction tuning (IT) of large language models (LLMs) has shown exceptional capability in understanding language across various tasks \cite{ouyang2022training}. However, the large number parameters of LLMs makes it challenging to transfer the pre-trained knowledge to downstream tasks \cite{naveed2023comprehensive, xu-etal-2024-team}. Alternatively, in-context learning (ICL) enables LLMs to perform downstream tasks by conditioning on task-specific prompts, thus eliminating the need for explicit gradient updates \cite{dong2022survey,wang2024learning, jiang2023team}. Recent successful deployment of LLMs in practical applications largely hinges on the effectiveness of the ICL and the IT.

Previous studies have extensively assessed the zero-shot capabilities of LLMs in computational social science (CSS) tasks, including hate speech detection \cite{roy2023probing} and rumour stance detection \cite{yang2024reinforcement}. However, CSS is a dynamic research area that involves detailed linguistic analysis and deep semantic comprehension. Direct zero-shot prompting LLMs to CSS tasks may even underperform compared to fully fine-tuned, task-specific smaller models like BERT \cite{juan2024fine}. Meanwhile, studies on ICL and IT typically occur independently, with direct comparisons between these approaches often overlooked.

To address the above issues, this paper raises the following research questions (\textbf{RQ}):

\begin{itemize}
\item \textbf{RQ 1:} What are the performance differences between LLMs with ICL and IT in few-shot CSS tasks?

\item \textbf{RQ 2:} How do varying numbers of sample influence the performance of LLMs with ICL and IT?

\item \textbf{RQ 3:} How different prompting strategies affect the proficiency of LLMs in CSS tasks?
\end{itemize}

To answer the above questions, we extensively investigate six open-source LLMs in a total of five publicly accessible social media datasets within n-shot settings, where $n \in \{1, 8, 16, 32\}$. 

Initially, we compare the few-shot classification performance of LLMs with ICL and IT separately. We then assess how performance varies with an increase in the number of samples. Lastly, we apply three prompting strategies including zero-shot, ICL and chain-of-thought (CoT), and examine their effects on performance. Additionally, except zero-shot setting, all experiments are conducted using five random seeds to account for the potential impact of few-shot sample quality on performance.

Our main findings are:

\begin{itemize}
\item[-] \textbf{(\uppercase\expandafter{\romannumeral1})} In few-shot settings, the performance of LLMs with ICL generally surpasses that of LLMs with IT on five CSS tasks.

\item[-] \textbf{(\uppercase\expandafter{\romannumeral2})} Merely increasing the sample size (from 1-shot to 32-shot in our experiments) does not consistently improve the performance of LLMs either with ICL or IT, and even leads to a performance decline in some cases.

\item[-] \textbf{(\uppercase\expandafter{\romannumeral3})} ICL prompting still outperforms zero-shot and CoT strategies, indicating that excessively complex prompting strategies can potentially hinder performance.
\end{itemize}

\section{Related Work}

\subsection{Instruction tuning for large language models}

IT \cite{jiang2023similarity, wang2024survey, parthasarathy2024ultimate} is an effective technique that updates LLM parameters in a supervised fashion by modifying the context of inputs to follow specific instructions. Previous studies have extensively discussed the advancements of IT in LLMs. For example, \citet{zhang2023instruction, qin2024unleashing} provide a comprehensive overview of the IT in LLMs, explaining the process of fine-tuning LLMs with instruction pairs and analyzing key factors that impact IT results. \citet{ouyang2022training} thoroughly examines data selection strategies for IT in LLMs, emphasizing the critical role of data quality over quantity and offering methods for selecting the most effective datasets to improve LLMs' instruction-following abilities. \citet{hu2024fine} proposes a Sequential Instruction Tuning (SIT) method that systematically incorporates continuous tasks into the training process to enhance the model’s capability to follow long, multi-step instructions. However, the aforementioned studies primarily assess IT in data-rich or zero-shot settings, leaving the few-shot performance of IT relatively underexplored.

\subsection{Comparison between instruction tuning and in-context learning}

ICL enables LLMs to quickly adapt to tasks by learning from samples without updating the model's weights \cite{yang2023not, brown2020language}. \citet{dong2022survey} comprehensively summarizes the progress and challenges of ICL, discussing related techniques including prompt design and training strategies, and explores effective application scenarios of ICL in enhancing the inferential capabilities of LLMs. \citet{coda2023meta} further explores how LLMs enhance their capabilities through the ICL paradigm by adjusting learning strategies and prior knowledge, through regression and multi-armed bandit tasks.

Recent studies have also focused on exploring the connections between IT and ICL. For example, \citet{mosbach2023few} evaluates the generalization capabilities of Pattern-based fine-tuning (PBFT) and ICT for out-of-domain (OOD) tasks under the same experimental settings in a few-shot context. They find that PBFT achieves better generalization. \citet{duan2023exploring} investigates how ICL and IT modify the hidden layer states of LLMs to achieve task adaptability in LLMs, finding that ICL is implicit IT. Our work differs from previous research in that we directly compare the classification performance between the ICL and IT in various CSS tasks.


\subsection{Large language models in computational social science}

LLMs have demonstrated exceptional capabilities in CSS \cite{moller2024prompt, jiang2023team, xu-etal-2024-team, jiang2023similarity}. For example, \citet{ziems2024can} has outlined a roadmap for using LLMs as tools for CSS, recommending best practices for prompting and conducting an extensive evaluation of the zero-shot performance of thirteen language models across twenty-four representative CSS benchmark tasks. Additionally, \citet{mu2024navigating} has assessed the zero-shot performance of two LLMs under six CSS tasks, while also researching the effects of various prompting strategies. However, the emerging CSS topics demand that LLMs quickly adapt to limited annotated data \cite{jiang2024cross}, therefore it is crucial to evaluate their few-shot performance in CSS tasks. Our work aims to explore the performance differences between ICL and IT in CSS tasks within few-shot settings, as well as how to enhance the capabilities of LLMs.

\begin{figure*}[htbp]
    \centering
    \includegraphics[width=1.0\textwidth]{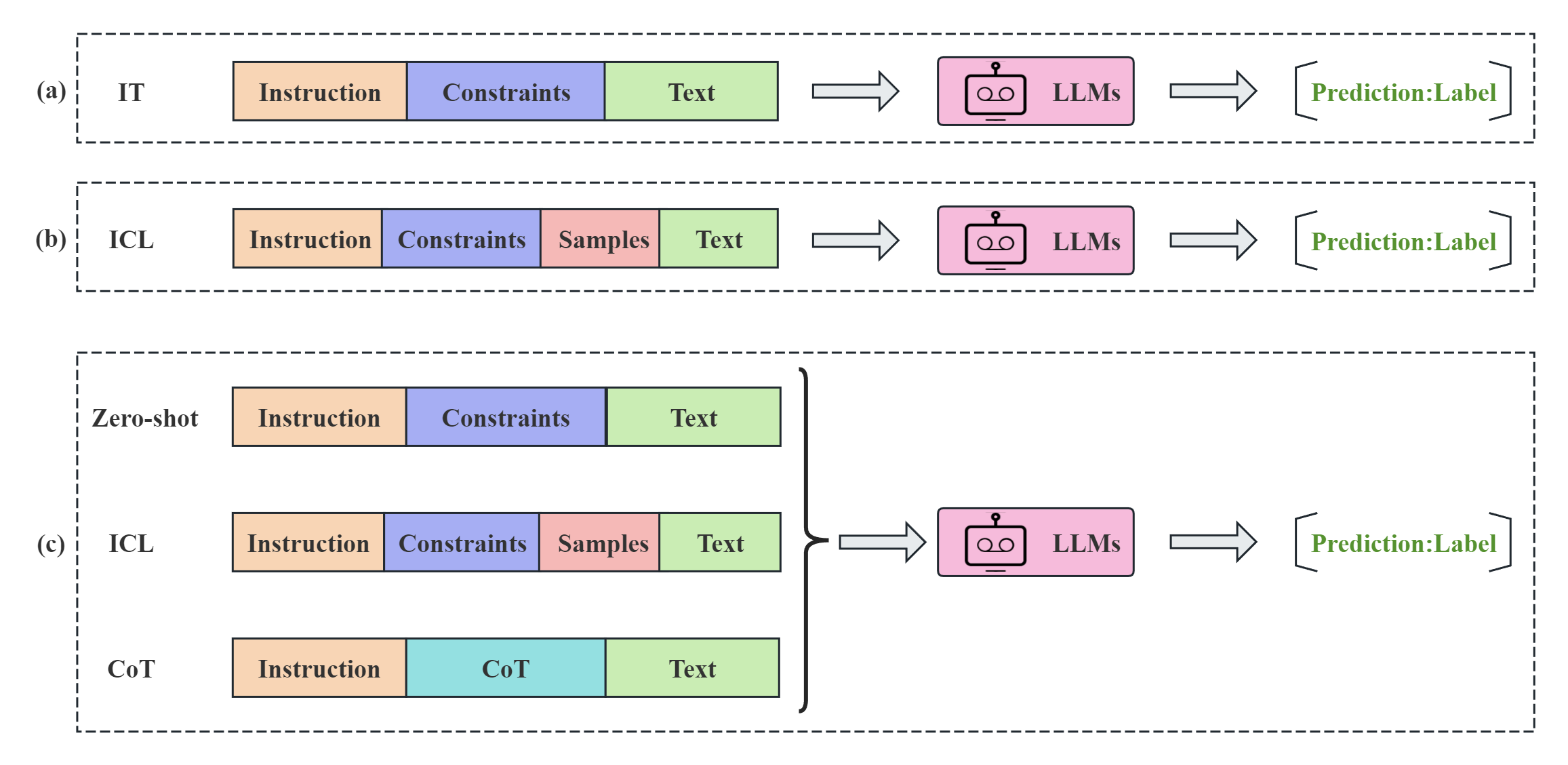} 
    \caption{Illustration of the overall workflow in this paper. (a) The instruction prompts including the context of the tasks (\textbf{Instruction}), the constraints for generating the responses from LLMs (\textbf{Constraints}), and the input text of each task (\textbf{Text}). (b) The ICL prompts include a set of input-label pairs (\textbf{Samples}) to guide the LLMs in focusing on task-specific content. (c) A comparison between different prompting strategies in CSS tasks.}
    \label{fig:workflow}
\end{figure*}

\section{Methodology}

\subsection{Instruction tuning for CSS}

Following the IT strategy outlined by \cite{duan2023exploring}, we first create a task-specific \textbf{Instruction} (e.g., "\textit{Analyze the content and determine if it includes ${label}$}", where ${label}$ represents task-specific labels) to define the objective of each task. We then incorporate a set of \textbf{Constraints} (e.g., "\textit{Respond only with ${label}$ or not ${label}$, without offering additional context or explanation}") to guide the LLMs' responses. The detailed workflow of the IT process is presented in Figure \ref{fig:workflow} (a) and Appendix \ref{sec:ins&con}.


Considering the computational efficiency and challenges of fine-tuning LLMs, we employ LoRA \cite{hu2021lora} for instruction-tuning across all models. Specifically, we set the dropout probability at 0.1 and the learning rate at 1e-4. As recommended by \citet{duan2023exploring}, the scaling factor is set to 32, with a rank of 8. The models are fine-tuned over three epochs using Brain Floating Point 16 (bf16) precision.


\subsection{In-context learning for CSS}
\label{sec:ICL}

In accordance with the in-context learning prompts described by \citet{jiang2024large}, we create input prompts consisting of \textbf{Instruction}, \textbf{Constraints}, \textbf{Samples} (e.g., "\textit{Tweet: How to not miss someone who doesn't even know you. Label: not bragging}"), and \textbf{Text} (e.g., "\textit{Tweet: For real, I just want to be prescribed something..., and what I'm all about. Label: }"). The detailed workflow of ICL is depicted in Figure \ref{fig:workflow} (b).

Given the limited fixed context length of LLMs, for the GossipCop dataset, we manually truncate the length of news articles to 256 tokens. Preliminary experiments revealed that higher temperature settings introduced more randomness in the model's responses. Hence, following the approach of \citet{mu2024navigating}, we apply a reduced temperature of 0.2 to enhance the model's focus and stability.

\subsection{Comparing in different prompting strategies}

In the zero-shot setting, we compose the prompt by combining \textbf{Instruction}, \textbf{Constraints}, and \textbf{Text}. For ICL, the detailed workflow is presented in Section \ref{sec:ICL}.

Inspired by \citet{dogan2024evaluating}, we utilize the ChatGPT-4 model\footnote{\url{https://chatgpt.com/}} to automatically generate \textbf{CoT} descriptions for each sample. For example, we input the tweet along with prompts in Bragging (e.g., ``\textit{Analyze the content and determine if it includes a bragging statement by using the CoT method. Tweet: For a minute I was tired of being the bigger man, until I realized that's just who I am''}). These \textbf{CoT} descriptions are then combined with \textbf{Instruction} and \textbf{Text} to form input prompts, as illustrated in Figure \ref{fig:workflow} (c). The examples of CoT description are provided in Appendix \ref{sec:cot}. 



%
\begin{table*}[htbp]
\centering
\begin{tabular}{l|c|cccc}
\hline
\textbf{Dataset} & \textbf{Total} & \multicolumn{4}{c}{\textbf{Labels (number of samples)}}                        \\ \hline
Bragging         & 6,696           & \multicolumn{2}{c}{Bragging (781)}    & \multicolumn{2}{c}{Not Bragging (5,915)}  \\ \hline
Complaint        & 3,449           & \multicolumn{2}{c}{Complaint (1,232)} & \multicolumn{2}{c}{Not Complaint (2,217)} \\ \hline
Sarcasm          & 4,868           & \multicolumn{2}{c}{Sarcasm (1,067)}   & \multicolumn{2}{c}{Not Sarcasm (3,801)}   \\ \hline
Rumour Stance    & 5,568           & Support(1,004) & Deny(415) & Query(464) & Comment(3,685) \\ \hline
GossipCop        & 6,805           & \multicolumn{2}{c}{Real (4,928)}     & \multicolumn{2}{c}{Fake (1,877)}          \\ \hline
\end{tabular}
\caption{Statistics of the selected datasets.}
\label{tab:1}
\end{table*}

\renewcommand{\arraystretch}{1.5}
\definecolor{myblue}{rgb}{0.8, 0.85, 0.95}

\begin{table*}[bp]
\centering
\small
\setlength{\tabcolsep}{4.3mm}{
\begin{tabular}{ccccccccccc}
\hline
\rowcolor{myblue}
\multicolumn{8}{c}{\textit{\textbf{1-Shot Setting}}}                                                                                                                                                                                 \\ \hline
\textbf{}    & \textbf{Qwen2}                    & \textbf{Baichuan2}                & \textbf{GLM4}                     & \textbf{Llama3}                   & \textbf{Gemma2}                   & \textbf{Phi-3}        & \textbf{Avg}            \\ \cline{2-8} 
\textbf{ICL} & \textbf{68.6}/\underline{67.2} & \textbf{61.6}/\underline{48.3} & \textbf{66.4}/\underline{60.0} & \textbf{60.1}/\underline{54.6} & 61.1/55.8          & \textbf{74.7}/\underline{62.5} & \textbf{65.4}/\underline{58.1}     \\
\textbf{IT}  & 65.1/62.2          & 60.7/47.9          & 56.0/51.3          & 53.9/50.1          & \textbf{71.3}/\underline{62.8} & 65.4/57.8    &62.1/55.3      
\\ \hline
\rowcolor{myblue}
\multicolumn{8}{c}{\textit{\textbf{8-Shot   Setting}}}                                                                                                                                                                               \\ \hline
\textbf{}    & \textbf{Qwen2}                    & \textbf{Baichuan2}                & \textbf{GLM4}                     & \textbf{Llama3}                   & \textbf{Gemma2}                   & \textbf{Phi-3}          & \textbf{Avg}            \\ \cline{2-8}
\textbf{ICL} & \textbf{71.7}/\underline{70.3} & 63.4/48.0          & \textbf{60.8}/\underline{56.6} & \textbf{61.6}/\underline{56.4} & 59.6/55.7          & \textbf{72.5}/\underline{62.4} & \textbf{64.9}/\underline{58.2}\\
\textbf{IT}  & 64.8/62.8          & \textbf{63.8}/\underline{49.3} & 56.3/51.5          & 53.0/50.0          & \textbf{68.8}/\underline{60.8} & 65.3/58.0   &62.0/55.4       \\ \hline
\rowcolor{myblue}
\multicolumn{8}{c}{\textit{\textbf{16-Shot   Setting}}}                                                                                                                                                                              \\ \hline
\textbf{}    & \textbf{Qwen2}                    & \textbf{Baichuan2}                & \textbf{GLM4}                     & \textbf{Llama3}                   & \textbf{Gemma2}                   & \textbf{Phi-3}                  & \textbf{Avg}            \\ \cline{2-8} 
\textbf{ICL} & \textbf{71.5}/\underline{70.1} & \textbf{62.6}/47.0 & \textbf{60.2}/\underline{56.7} & \textbf{60.4}/\underline{55.5} & 60.8/56.7          & \textbf{70.6}/\underline{61.5} & \textbf{64.4}/\underline{57.9}\\
\textbf{IT}  & 64.4/62.7          & 62.3/\underline{49.5}          & 56.3/51.4          & 51.2/48.6          & \textbf{68.0}/\underline{59.9} & 65.1/57.5    &61.2/54.9      \\ \hline
\rowcolor{myblue}
\multicolumn{8}{c}{\textit{\textbf{32-Shot   Setting}}}                                                                                                                                                                              \\ \hline
\textbf{}    & \textbf{Qwen2}                    & \textbf{Baichuan2}                & \textbf{GLM4}                     & \textbf{Llama3}                   & \textbf{Gemma2}                   & \textbf{Phi-3}                   & \textbf{Avg}            \\ \cline{2-8}
\textbf{ICL} & \textbf{72.6}/\underline{71.2} & \textbf{70.2}/\underline{50.2} & \textbf{61.7}/\underline{57.6} & \textbf{59.6}/\underline{55.1} & 61.0/56.4          & \textbf{69.4}/\underline{59.2} & \textbf{65.7}/\underline{58.3}\\
\textbf{IT}  & 65.9/63.4          & 61.3/48.5          & 56.2/51.4          & 52.4/49.4          & \textbf{71.4}/\underline{62.2} & 64.8/57.1   & 62.0/55.3       \\ \hline
\end{tabular}%
}
\caption{The LLMs' performance is compared between ICL and IT in CSS tasks. Scores are first calculated by averaging the accuracy and macro-F1 (Acc/F1) scores (\%) across five seeds for each model, followed by computing the mean across five tasks. \textbf{Bold} indicates the highest accuracy, and \underline{Underline} denotes the best F1 score.}
\label{tab:2}
\end{table*}

\section{Experimental setups}

\subsection{Data}

To assess the classification performance of LLMs, five publicly available datasets are selected, encompassing a broad spectrum of computational social science topics. The statistics of these datasets are presented in Table \ref{tab:1}.

\textbf{Bragging} \cite{jin2022automatic} \textbf{:} This dataset is designed to facilitate a comprehensive semantic analysis of tweets to ascertain whether they contain narratives of bragging, specifically identifying the subject of the author's boast.

\textbf{Complaint} \cite{preoctiuc2019automatically} \textbf{:} This task aims to identify whether tweets from social media contain complaints, where the complaint content expresses a mismatch between reality and expectations in a specific context.

\textbf{Sarcasm} \cite{farha2022semeval} \textbf{:} This task aims to conduct semantic analysis on texts to determine whether they contain sarcasm.

\textbf{Rumour Stance} \cite{derczynski2017semeval} \textbf{:} This task aims to perform semantic analysis on tweets (rumours) in social media to assess the stance classification of the rumours.

\textbf{GossipCop} \cite{shu2020fakenewsnet} \textbf{:} This task aims to perform semantic analysis on news articles in entertainment media to determine the authenticity of the news articles. 


For each benchmark task, we utilize stratified random sampling to divide the dataset into 70\% for training, 10\% for validation, and 20\% for testing. The 10\% validation set is used for hyperparameter tuning during the instruction tuning, and the performances of LLMs under the ICL and IT are evaluated on the designated 20\% test set.

We apply the same few-shot settings to both the ICL and IT. First, we randomly sample $ n\in \{1, 8, 16, 32\}$ examples (where $n$ is the number of samples per class) from the training set. Given the high sensitivity of ICL and IT to the choice of examples, we use five random seeds per shot, repeating the process to assess LLM performance in few-shot scenarios.

\subsection{Baselines}


To ensure a fair comparison of LLMs in CSS tasks, we utilize Huggingface\footnote{\url{https://huggingface.co/}} to select six different open-source LLMs, with model sizes ranging from 7B to 9B, namely Qwen2-7B-Instruct (Qwen2) \cite{yang2024qwen2}, Baichuan2-7B-Chat (Baichuan2) \cite{yang2023baichuan}, GLM4-9B-chat (GLM4) \cite{glm2024chatglm}, Meta-llama3-8B-instruct (LLama3) \cite{meta2024introducing}, Gemma-2-9B-it (Gemma2) \cite{team2024gemma}, and Phi-3-Small-128K-Instruct (Phi-3) \cite{abdin2024phi}.

\section{Results}


The overall experimental results are presented in Table \ref{tab:2} and Table \ref{tab:3}. For each n-shot setting, we evaluate the LLMs by computing the average accuracy (Acc) and macro-F1 (F1) scores across five random seeds\footnote{The detailed experimental results for each seed are presented in Appendix \ref{sec:results}}.

\begin{figure*}[bp]
\centering
  \begin{subfigure}{0.5\linewidth}
    \includegraphics[width=\linewidth]{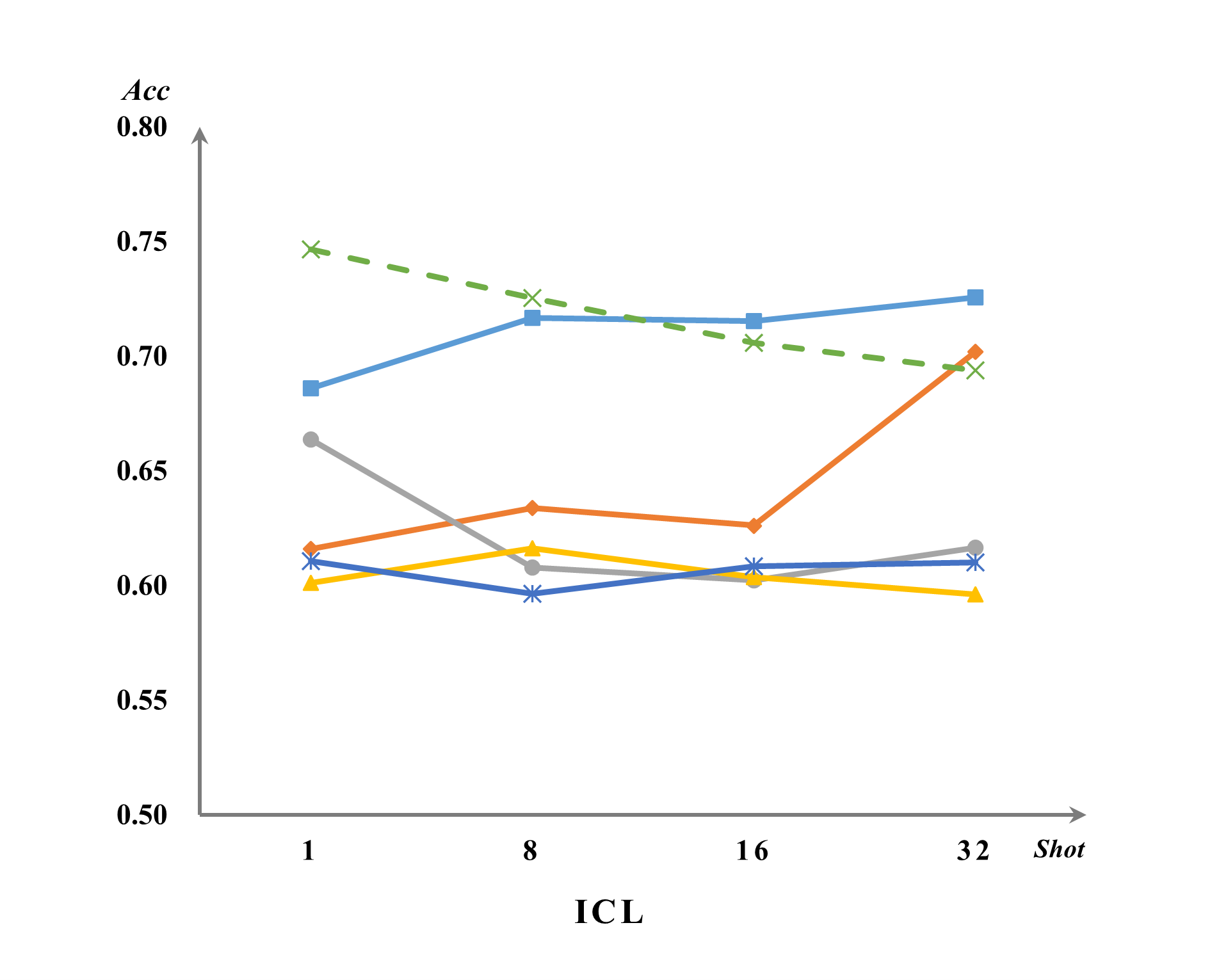}
    \label{fig:icl}
  \end{subfigure}%
  \begin{subfigure}{0.5\linewidth}
    \includegraphics[width=\linewidth]{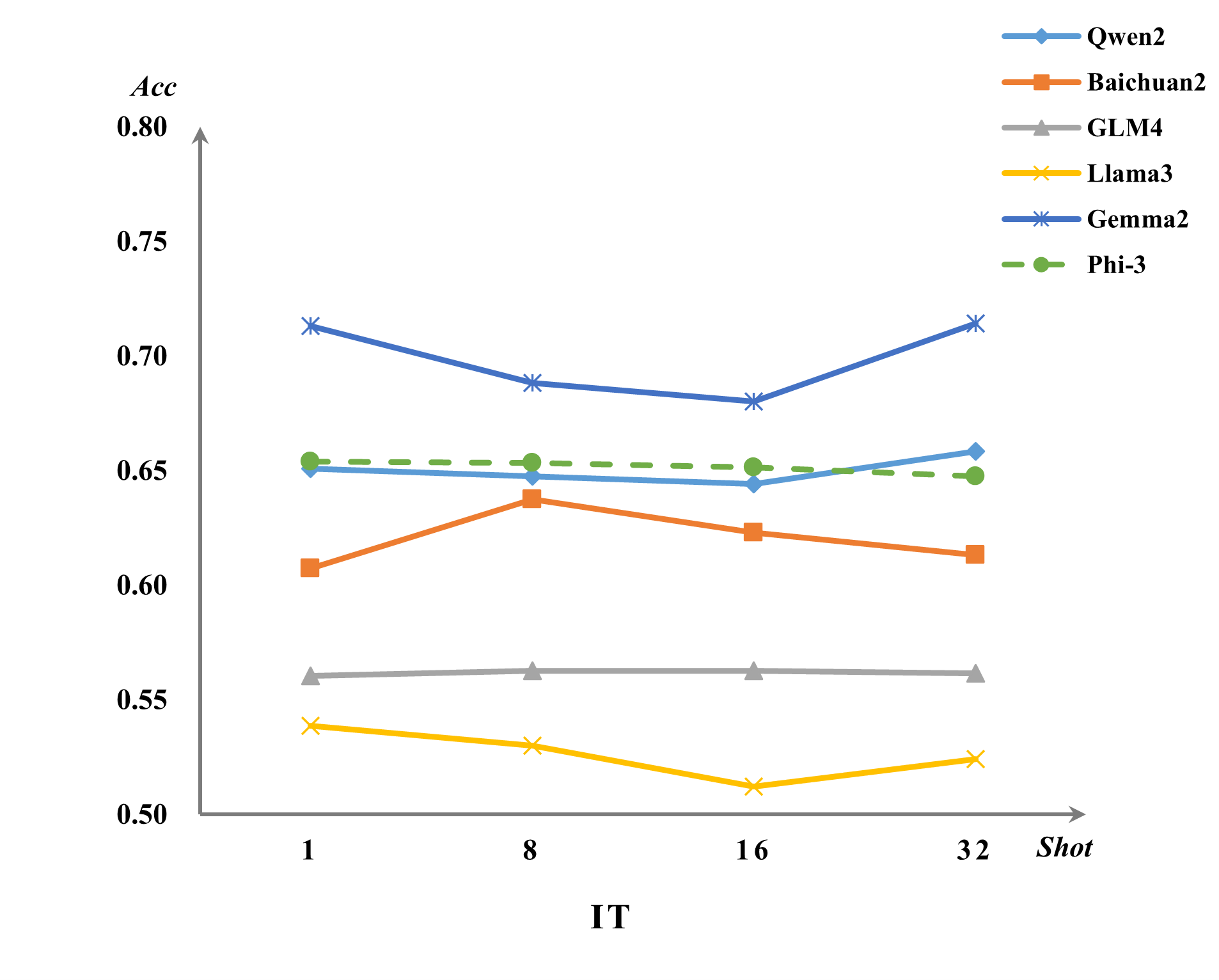}
    \label{fig:it}
  \end{subfigure}
  \caption{Illustration of different sample sizes affect the performance of LLMs with ICL and IT respectively.}
  \label{fig:both_images}
\end{figure*}

\renewcommand{\arraystretch}{1.5}
\definecolor{myblue}{rgb}{0.8, 0.85, 0.95}

\begin{table*}[]
\small
\setlength{\tabcolsep}{1.3mm}{

\begin{tabular}{lcccccccccc}
\hline
\rowcolor{myblue}
\multicolumn{11}{c}{\textit{\textbf{1-Shot Setting}}}                                                                                                                                                                                                                                   \\
\hline
\multirow{2}{*}{\textbf{Model}} & \multicolumn{2}{c}{\textbf{Bragging}}           & \multicolumn{2}{c}{\textbf{Complaint}}          & \multicolumn{2}{c}{\textbf{Sarcasm}}            & \multicolumn{2}{c}{\textbf{Rumour Stance}}      & \multicolumn{2}{c}{\textbf{GossipCop}}          \\ \cline{2-11}
                                & \textbf{ICL}       & \textbf{IT}        & \textbf{ICL}       & \textbf{IT}        & \textbf{ICL}       & \textbf{IT}        & \textbf{ICL}       & \textbf{IT}        & \textbf{ICL}       & \textbf{IT}        \\
                                \hline
\textbf{Qwen2}                  & 86.0 / \underline{86.9}   & 81.7/84.0            & 81.5/81.8          & 85.6/85.9 & 59.3/62.5 & 51.4/54.1          & 41.3/\underline{44.0}   & 33.2/27.9          & \textbf{75.0}/60.6   & 73.5/59.1          \\
\textbf{Baichuan2}              & 81.7/60.8          & 85.7/58.8 & 62.6/60.7 & 62.0/58.1            & \textbf{77.7}/50.9 & 55.4/43.0            & 48.4/32.8 & 41.2/29.8          & 37.6/36.4          & 59.2/50.0   \\
\textbf{GLM4}                   & 87.1/73.7 & 58.6/51.8          & 83.4/82.9 & 79.8/79.4          & 54.8/53.4 & 53.5/51.3          & 36.1/28.3 & 17.9/16.0            & 70.5/61.7 & 70.4/57.8          \\
\textbf{Llama3}                 & 78.7/64.6 & 67.7/56.7          & 88.4/87.8 & 81.8/81.3          & 43.4/43.2 & 35.1/34.3          & 22.5/18.6 & 18.6/20.2          & 67.6/58.8 & 66.2/57.9          \\
\textbf{Gemma2}                 & 77.4/64.3          & 84.9/70.0   & 84.0/83.6            & 85.5/85.0   & 41.4/41.3          & 57.2/55.9 & 45.5/35.5          & 55.7/40.7 & 57.1/54.5          & 73.2/\underline{62.3} \\
\textbf{Phi-3}                  & \textbf{89.1}/71.0    & 87.8/72.2          & \textbf{89.6}/\underline{88.6} & 86.3/85.7          & 68.2/\underline{63.3} & 43.9/43.8          & \textbf{57.6}/41.1 & 38.2/35.2          & 68.8/48.5          & 70.8/52.3 \\

\hline
\rowcolor{myblue}
\multicolumn{11}{c}{\textit{\textbf{8-Shot Setting}}}                                                                                                                                                                                                                                     \\
\hline
\multirow{2}{*}{\textbf{Model}} & \multicolumn{2}{c}{\textbf{Bragging}}           & \multicolumn{2}{c}{\textbf{Complaint}}          & \multicolumn{2}{c}{\textbf{Sarcasm}}            & \multicolumn{2}{c}{\textbf{Rumour Stance}}      & \multicolumn{2}{c}{\textbf{GossipCop}}          \\ \cline{2-11}
                                & \textbf{ICL}       & \textbf{IT}        & \textbf{ICL}       & \textbf{IT}        & \textbf{ICL}       & \textbf{IT}        & \textbf{ICL}       & \textbf{IT}        & \textbf{ICL}       & \textbf{IT}        \\
                                \hline
\textbf{Qwen2}                  & 86.3/\underline{87.3} & 77.7/81.1          & 89.2/\underline{89.3} & 84.6/84.9          & 70.8/\underline{72.6} & 52.7/55.8          & 35.5/39.1          & 38.4/30.6 & \textbf{76.5}/63.1 & 70.4/61.7          \\
\textbf{Baichuan2}              & 83.1/65.2 & 79.0/59.6            & 64.1/61.5 & 61.0/53.2            & \textbf{78.2}/44.3 & 60.7/52.9          & 51.7/29.5 & 48.6/28.4          & 39.8/39.6          & 69.4/52.3 \\
\textbf{GLM4}                   & 80.2/68.4 & 59.0/52.1            & 84.4/84.0   & 79.9/79.5          & 43.2/43.2          & 54.1/51.9 & 24.0/24.0     & 18.0/16.0              & 72.1/\underline{63.7} & 70.3/57.8          \\
\textbf{Llama3}                 & 83.3/68.6 & 63.4/54.0            & 88.1/86.6 & 81.3/80.8          & 50.8/50.2 & 38.8/38.4          & 26.3/20.4          & 18.7/19.8          & 59.6/56.4          & 63.0/56.8   \\
\textbf{Gemma2}                 & 72.3/61.5          & 81.8/67.4 & 85.1/84.7 & 83.9/83.5          & 42.5/42.5          & 57.0/55.7   & 45.9/38.1          & \textbf{56.8}/\underline{40.9}  & 52.4/51.6          & 64.7/56.7 \\
\textbf{Phi-3}                  & \textbf{88.9}/73.8 & 87.9/72.7          & \textbf{89.7}/88.9 & 86.3/85.7          & 62.6/59.5 & 44.2/44.1          & 51.8/38.3 & 37.2/34.6          & 69.7/51.5          & 71.1/52.7 \\

\hline
\rowcolor{myblue}
\multicolumn{11}{c}{\textit{\textbf{16-Shot Setting}}}                                                                                                                                                                                                                                              \\

\hline
\multirow{2}{*}{\textbf{Model}} & \multicolumn{2}{c}{\textbf{Bragging}}           & \multicolumn{2}{c}{\textbf{Complaint}}          & \multicolumn{2}{c}{\textbf{Sarcasm}}            & \multicolumn{2}{c}{\textbf{Rumour Stance}}      & \multicolumn{2}{c}{\textbf{GossipCop}}          \\ \cline{2-11}
                                & \textbf{ICL}       & \textbf{IT}        & \textbf{ICL}       & \textbf{IT}        & \textbf{ICL}       & \textbf{IT}        & \textbf{ICL}       & \textbf{IT}        & \textbf{ICL}       & \textbf{IT}        \\
                                \hline
\textbf{Qwen2}                  & 84.9/\underline{86.4} & 76.8/80.5          & 89.4/\underline{89.5} & 84.1/84.4          & 71.5/\underline{73.1} & 53.0/56.1            & 35.9/39.6          & 38.6/30.6 & \textbf{76.1}/61.9 & 69.8/61.8          \\
\textbf{Baichuan2}              & 84.5/67.4 & 83.8/62.7          & 63.6/57.3 & 62.0/53.5            & \textbf{78.2}/44.2 & 61.5/52.6          & 47.0/26.9   & 44.6/29.6          & 39.8/39.5          & 59.6/48.8 \\
\textbf{GLM4}                   & 81.1/69.2 & 58.8/51.9         & 84.1/83.8 & 79.8/79.5          & 42.1/42.0            & 54.1/51.8 & 23.1/24.1 & 18.1/16.0            & 70.8/\underline{64.3} & 70.4/57.9          \\
\textbf{Llama3}                 & 83.9/68.8 & 66.4/55.9          & 86.6/84.4 & 80.7/80.3          & 52.9/52.1 & 36.6/36.3          & 21.9/17.8 & 18.6/19.2          & 56.5/54.5 & 53.8/51.3          \\
\textbf{Gemma2}                 & 68.6/58.6          & 82.7/68.2 & 86.6/86.1 & 83.8/83.4          & 48.2/48.0            & 56.5/55.2 & 46.3/37.4          & \textbf{56.6}/\underline{40.6} & 54.6/53.6     &     60.5/51.8 \\
\textbf{Phi-3}                  & \textbf{88.1}/73.7 & 87.6/72.1          & \textbf{89.6}/88.8 & 85.4/84.8          & 54.8/53.6 & 44.4/44.3          & 50.2/40.4 & 37.9/35.1          & 70.2/50.8          & 70.4/51.3 \\
\hline
\rowcolor{myblue}
\multicolumn{11}{c}{\textit{\textbf{32-Shot Setting}}}                                                                                                                                                                                                                                    \\
\hline
\multirow{2}{*}{\textbf{Model}} & \multicolumn{2}{c}{\textbf{Bragging}}           & \multicolumn{2}{c}{\textbf{Complaint}}          & \multicolumn{2}{c}{\textbf{Sarcasm}}            & \multicolumn{2}{c}{\textbf{Rumour Stance}}      & \multicolumn{2}{c}{\textbf{GossipCop}}          \\ \cline{2-11}
                                & \textbf{ICL}       & \textbf{IT}        & \textbf{ICL}       & \textbf{IT}        & \textbf{ICL}       & \textbf{IT}        & \textbf{ICL}       & \textbf{IT}        & \textbf{ICL}       & \textbf{IT}        \\
                                \hline

\textbf{Qwen2}                  & 86.7/\underline{87.7} & 76.2/80.1          & \textbf{89.9}/\underline{90.0}   & 84.1/84.5          & 71.1/\underline{72.7} & 51.7/54.7          & 39.4/\underline{43.4}          & 46.6/36.3 & \textbf{75.8}/62.1 & 70.6/61.6          \\
\textbf{Baichuan2}              & 87.1/67.5 & 80.8/60.7          & 73.7/67.8 & 62.4/53.5          & \textbf{78.2}/44.2 & 61.0/53.9            & \textbf{65.7}/26.0   & 44.4/29.5          & 46.2/45.4          & 58.0/44.8   \\
\textbf{GLM4}                   & 81.4/69.4 & 58.8/51.9          & 84.4/84.0   & 79.8/79.5          & 42.1/42.1          & 53.9/51.7 & 29.2/28.0   & 17.9/16.0            & 71.2/\underline{64.7} & 70.4/57.8          \\
\textbf{Llama3}                 & 84.5/69.5 & 64.5/54.8          & 87.1/85.1 & 81.6/81.1          & 48.9/48.6 & 36.4/36.0            & 17.3/15.2 & 26.1/23.8          & 60.3/57.0   & 53.6/51.1          \\
\textbf{Gemma2}                 & 70.3/59.7          & 83.1/68.4 & 87.0/86.5   & 84.0/83.6            & 53.8/53.1          & 56.0/54.8   & 45.9/35.0            & 60.3/42.7 & 48.0/47.7            & 73.8/61.5 \\
\textbf{Phi-3}                  & 86.3/72.3          & \textbf{87.7}/72.3 & 89.0/88.2   & 85.0/84.4            & 49.3/48.9 & 44.0/43.9            & 51.1/36.9 & 36.9/34.5          & 71.2/49.8 & 70.2/50.5 \\

\hline

\end{tabular}}
\caption{The accuracy and macro-F1 (Acc/F1) scores (\%) of various LLMs across five benchmark tasks. For each shot, experiments are conducted using five random seeds, and the average values across all seeds are recorded. In each task, \textbf{Bold} indicates the highest accuracy, and \underline{Underline} is the best F1 score.}
\label{tab:3}
\end{table*}

\renewcommand{\arraystretch}{1.5}
\begin{table*}[t]
\centering
\small
\setlength{\tabcolsep}{3.7mm}{
\begin{tabular}{cccccccc}
\hline
                  & \textbf{Qwen2} & \textbf{Baichuan2} & \textbf{GLM4} & \textbf{Llama3} & \textbf{Gemma2} & \textbf{Phi-3} & \textbf{Avg} \\\hline
\textbf{Zero-shot} & 61.7/55.9      & 55.0/44.0              & 58.1/54.6     & 58.8/52.5       & \textbf{64.3}/\underline{57.1}       & 64.5/54.7      & 60.4/53.1    \\
\textbf{ICL}      & \textbf{68.6}/\underline{67.2}      & \textbf{61.6}/48.3          & \textbf{66.4}/\underline{60.0}       & \textbf{60.1}/\underline{54.6}       & 61.1/55.8       & \textbf{74.7}/\underline{62.5}      & \textbf{65.4}/\underline{58.1}    \\
\textbf{CoT}      & 66.6/63.8      & 56.7/\underline{52.0}          & 65.6/\underline{60.0}     & 59.4/52.6       & 53.8/52.8       & 67.0/62.1        & 61.5/57.2 \\\hline  
\end{tabular}
}
\caption{The accuracy and macro-F1 (Acc/F1) scores(\%) across all benchmark tasks for different models represent the average values from five tasks. \textbf{Bold} and \underline{Underline} indicate the highest accuracy and F1 among the LLMs in each task respectively.}
\label{tab:4}
\end{table*}



\textbf{Comparing between IT and ICL: } We first calculate the average accuracy and F1 scores across five seeds for each model, and then compute the means of these scores across all CSS tasks. The averaged scores are presented in Table \ref{tab:2}. We observe that the overall classification performance of LLMs with ICL is significantly better than that of LLMs with IT. For instance, ICL outperforms IT by 3.3\% in accuracy in the 1-shot setting. Similarly, LLMs with ICL consistently outperform LLMs with IT in the 8, 16, and 32-shot settings, with accuracy improvements of 2.9\%, 3.2\%, and 3.7\%, respectively.


We also examine how different tasks impact the performance of ICL and IT, as presented in Table \ref{tab:3}. In the Bragging and Complaint tasks, ICL consistently outperforms IT, achieving higher accuracy across all six models. For instance, ICL attains an average accuracy of 85.2\% across six LLMs, which is 5.7\% higher than IT in the 32-shot setting for the Complaint task. This advantage is also evident in GossipCop, Sarcasm, and Rumour Stance. However, it is noteworthy that LLMs demonstrate relatively lower performance in the latter two benchmark tasks (e.g., Sarcasm and Rumour Stance) compared to others. For example, the average accuracy of ICL in Sarcasm and Rumour Stance is 57.2\% and 41.4\%, respectively, which is significantly lower than the 85.2\% achieved in the Complaint task under the 32-shot setting. 



\textbf{Comparing between LLMs:} We also assess the ability of six LLMs to address CSS tasks using ICL and IT in few-shot settings, as presented in Table \ref{tab:3}. We observe that Phi-3 outperforms the others in most tasks, achieving the highest average accuracy in the Bragging and Complaint tasks, with scores of 88.1\% and 89.5\%, respectively. Baichuan2 and Qwen2 attain the highest average accuracy of 78.1\% and 75.9\% in the Sarcasm and GossipCop tasks, respectively. However, GLM4 and Llama3 generally underperform compared to the others. Additionally, all LLMs exhibit significant weaknesses in Rumour Stance. Notably, the IT performance of Gemma2 consistently surpasses that of ICL across all tasks. For instance, in the 1-shot setting, the average accuracy of IT exceeds that of ICL by 10.2\%.

\textbf{Comparing between different n-shot settings: } Figure \ref{fig:both_images} illustrates the overall performance of the LLMs with ICL and IT in different n-shot settings. We compute the average accuracy of the six LLMs across all tasks. The experimental results show that the performance of LLMs with either ICL or IT does not consistently improve as the number of training samples increases, and even declines in some cases. For example, the average accuracy of Phi-3 with ICL is 74.7\% in the 1-shot setting, but drops to 69.4\% in the 32-shot setting. Similarly, the accuracy of Llama3 with IT decreases from 53.9\% in the 1-shot setting to 52.4\% in the 32-shot setting.



\textbf{Comparing between different prompt strategies:} To assess the impact of prompting strategies on the inferential capabilities of LLMs, we compare three prompting approaches: zero-shot, ICL, and CoT in the 1-shot setting. Note that this comparison is not conducted in other n-shot settings, as we found no strong correlation between the number of samples and prompting strategies, based on the preliminary findings.

The performance of these three prompting strategies is shown in Table \ref{tab:4}. We observe that ICL prompting consistently achieves the highest accuracy and F1 scores. Specifically, ICL surpasses CoT in accuracy by 3.9\%. CoT, in turn, outperforms zero-shot by 1.1\% in accuracy. Lastly, zero-shot exhibits the lowest accuracy and F1 scores.

\section{Analysis}

The experimental results underscore the proficiency of LLMs in CSS tasks that require comprehension of complex real-world contexts. Next, we will contextualize these findings within the framework of our three research questions:

\noindent \textbf{(RQ1) What are the performance differences between LLMs with ICL and IT in few-shot CSS tasks?}

The experimental results reveal that LLMs with ICL generally outperform those with IT in few-shot CSS tasks. ICL exhibits strong adaptability, likely due to the extensive knowledge acquired during the pre-training phase. This allows the model to comprehend and swiftly adapt to complex tasks by leveraging pre-trained knowledge. While IT also enhances LLM capabilities through instructions, its performance is comparatively more sensitive than that of ICL, as illustrated in Figure \ref{fig:both_images}. 

Additionally, ICL enables the model to directly leverage the input-label pairs provided in the samples to guide inference without requiring gradient updates. For LLMs with IT, insufficient training samples can lead to overfitting, instability, and may even impair the inferential capacity of the models. For example, the average accuracies of GLM4 and Llama3 are 56.0\% and 53.9\% in the 1-shot setting. However, both models achieve higher average accuracies of 58.1\% and 58.8\% in the zero-shot settings, respectively.




\noindent \textbf{(RQ2) How do varying numbers of sample influence the performance of LLMs with ICL and IT?}

Our experimental results suggest that merely increasing the number of training samples does not consistently improve the performance of LLMs with either ICL or IT, and in some cases, it even leads to a decline.


Given the characteristics of few-shot settings, we speculate that the contextual diversity of samples is more crucial than their quantity, regardless of whether LLMs use IT or ICL. If the additional samples are highly similar in content, LLMs may struggle to learn from the feature diversity in few-shot examples, leading to poor inferential performance. Moreover, when the feature distribution of few-shot samples deviates significantly from that of the pre-trained data, this variation may also negatively affect the classification performance of LLMs.




\noindent \textbf{(RQ3) How different prompting strategies affect the proficiency of LLMs in CSS tasks?}

LLMs with ICL achieve the highest performance among the three prompting strategies: zero-shot, ICL, and CoT. This indicates that incorporating a small number of input-label pairs into the prompt can help LLMs better focus on task-specific content across various CSS tasks.

Surprisingly, we find that the CoT strategy slightly underperforms compared to ICL. We hypothesize two potential reasons for this: 1) the CoT examples are automatically generated by GPT-4, which may result in varying content quality depending on the context; 2) incorporating CoT descriptions into the prompt might introduce noise and transform a simple classification problem into a more complex language understanding task, as detailed in the Appendix \ref{sec:cot}.



Finally, the zero-shot strategy yields the lowest performance. This may be due to insufficient contextual information to guide the model in understanding and performing CSS tasks, which often require deeper semantic comprehension (e.g., Sarcasm and Bragging). Moreover, the zero-shot strategy primarily depends on the pre-trained knowledge of LLMs. The absence of task-specific knowledge during the pre-training phase may cause the model to struggle in identifying appropriate solutions.

\section{Conclusion}

In this paper, we first evaluate the performance of LLMs with IT and ICL in few-shot CSS tasks. We also investigate whether increasing the number of training samples affects LLM performance. Lastly, we compare different prompting strategies and analyze their efficiency in few-shot settings.

In our experiments, we evaluate six open-source LLMs on five publicly available CSS datasets. Our results indicate that: 1) LLMs with ICL generally outperform those with IT in tackling complex CSS tasks in few-shot settings; 2) merely increasing the number of samples without considering their quality does not consistently improve the performance of LLMs with either ICL or IT, and may even lead to a decline in some cases; 3) LLMs with ICL are more effective than those using zero-shot and CoT strategies in few-shot settings, suggesting that overly complex prompting may negatively impact LLM performance.

Overall, our research underscores the substantial advantages of ICL in handling CSS tasks in few-shot settings, highlighting the critical role of optimizing sample quality and prompting strategies to enhance the classification performance of LLMs.


\section{Limitations}


This study acknowledges several limitations, including: 1) Due to computational resource constraints and the context length limitations of LLMs, larger n-shot settings remain underexplored. 2) Our experiments primarily compare LLMs with parameters ranging between 7B and 9B, due to hardware restrictions. 3) The generation of CoT descriptions relies mainly on GPT-4, without manual assessment, which may result in inconsistencies in CoT quality.

\section{Ethic statement}

This work has received ethical approval from the Ethics Committee of our university and adheres to the research policies of Twitter. All datasets were obtained via links provided in the respective research papers or directly from the authors upon request. Additionally, we confirm that the data was fully anonymized prior to being used for model inference with the LLMs.
Due to the time-intensive and challenging nature of generating the CoT strategy, we solely rely on ChatGPT-4 to automatically generate the CoT descriptions, without manually crafting any CoT strategies.

\section*{Acknowledgments}
This work is funded by the Natural Science Foundation of Shandong Province under grant ZR2023QF151 and the Natural Science Foundation of China under grant 12303103.


\bibliography{custom}

\begin{thebibliography}{38}
\providecommand{\natexlab}[1]{#1}

\bibitem[{Abdin et~al.(2024)Abdin, Jacobs, Awan, Aneja, Awadallah, Awadalla, Bach, Bahree, Bakhtiari, Behl et~al.}]{abdin2024phi}
Marah Abdin, Sam~Ade Jacobs, Ammar~Ahmad Awan, Jyoti Aneja, Ahmed Awadallah, Hany Awadalla, Nguyen Bach, Amit Bahree, Arash Bakhtiari, Harkirat Behl, et~al. 2024.
\newblock Phi-3 technical report: A highly capable language model locally on your phone.
\newblock \emph{arXiv preprint arXiv:2404.14219}.

\bibitem[{Brown et~al.(2020)Brown, Mann, Ryder, Subbiah, Kaplan, Dhariwal, Neelakantan, Shyam, Sastry, Askell et~al.}]{brown2020language}
Tom Brown, Benjamin Mann, Nick Ryder, Melanie Subbiah, Jared~D Kaplan, Prafulla Dhariwal, Arvind Neelakantan, Pranav Shyam, Girish Sastry, Amanda Askell, et~al. 2020.
\newblock Language models are few-shot learners.
\newblock \emph{Advances in neural information processing systems}, 33:1877--1901.

\bibitem[{Coda-Forno et~al.(2023)Coda-Forno, Binz, Akata, Botvinick, Wang, and Schulz}]{coda2023meta}
Julian Coda-Forno, Marcel Binz, Zeynep Akata, Matt Botvinick, Jane Wang, and Eric Schulz. 2023.
\newblock Meta-in-context learning in large language models.
\newblock \emph{Advances in Neural Information Processing Systems}, 36:65189--65201.

\bibitem[{Derczynski et~al.(2017)Derczynski, Bontcheva, Liakata, Procter, Hoi, and Zubiaga}]{derczynski2017semeval}
Leon Derczynski, Kalina Bontcheva, Maria Liakata, Rob Procter, Geraldine Wong~Sak Hoi, and Arkaitz Zubiaga. 2017.
\newblock Semeval-2017 task 8: Rumoureval: Determining rumour veracity and support for rumours.
\newblock In \emph{Proceedings of the 11th International Workshop on Semantic Evaluation (SemEval-2017)}, pages 69--76.

\bibitem[{Dogan et~al.(2024)Dogan, Kesen, Calixto, Erdem, and Erdem}]{dogan2024evaluating}
Mustafa Dogan, Ilker Kesen, Iacer Calixto, Aykut Erdem, and Erkut Erdem. 2024.
\newblock Evaluating linguistic capabilities of multimodal llms in the lens of few-shot learning.
\newblock \emph{arXiv preprint arXiv:2407.12498}.

\bibitem[{Dong et~al.(2022)Dong, Li, Dai, Zheng, Wu, Chang, Sun, Xu, and Sui}]{dong2022survey}
Qingxiu Dong, Lei Li, Damai Dai, Ce~Zheng, Zhiyong Wu, Baobao Chang, Xu~Sun, Jingjing Xu, and Zhifang Sui. 2022.
\newblock A survey on in-context learning.
\newblock \emph{arXiv preprint arXiv:2301.00234}.

\bibitem[{Duan et~al.(2023)Duan, Tang, Yang, Abbasi, and Tam}]{duan2023exploring}
Hanyu Duan, Yixuan Tang, Yi~Yang, Ahmed Abbasi, and Kar~Yan Tam. 2023.
\newblock Exploring the relationship between in-context learning and instruction tuning.
\newblock \emph{arXiv preprint arXiv:2311.10367}.

\bibitem[{Farha et~al.(2022)Farha, Oprea, Wilson, and Magdy}]{farha2022semeval}
Ibrahim~Abu Farha, Silviu Oprea, Steve Wilson, and Walid Magdy. 2022.
\newblock Semeval-2022 task 6: isarcasmeval, intended sarcasm detection in english and arabic.
\newblock In \emph{The 16th International Workshop on Semantic Evaluation 2022}, pages 802--814. Association for Computational Linguistics.

\bibitem[{GLM et~al.(2024)GLM, Zeng, Xu, Wang, Zhang, Yin, Rojas, Feng, Zhao, Lai, Yu, Wang, Sun, Zhang, Cheng, Gui, Tang, Zhang, Li, Zhao, Wu, Zhong, Liu, Huang, Zhang, Zheng, Lu, Duan, Zhang, Cao, Yang, Tam, Zhao, Liu, Xia, Zhang, Gu, Lv, Liu, Liu, Yang, Song, Zhang, An, Xu, Niu, Yang, Li, Bai, Dong, Qi, Wang, Yang, Du, Hou, and Wang}]{glm2024chatglm}
Team GLM, Aohan Zeng, Bin Xu, Bowen Wang, Chenhui Zhang, Da~Yin, Diego Rojas, Guanyu Feng, Hanlin Zhao, Hanyu Lai, Hao Yu, Hongning Wang, Jiadai Sun, Jiajie Zhang, Jiale Cheng, Jiayi Gui, Jie Tang, Jing Zhang, Juanzi Li, Lei Zhao, Lindong Wu, Lucen Zhong, Mingdao Liu, Minlie Huang, Peng Zhang, Qinkai Zheng, Rui Lu, Shuaiqi Duan, Shudan Zhang, Shulin Cao, Shuxun Yang, Weng~Lam Tam, Wenyi Zhao, Xiao Liu, Xiao Xia, Xiaohan Zhang, Xiaotao Gu, Xin Lv, Xinghan Liu, Xinyi Liu, Xinyue Yang, Xixuan Song, Xunkai Zhang, Yifan An, Yifan Xu, Yilin Niu, Yuantao Yang, Yueyan Li, Yushi Bai, Yuxiao Dong, Zehan Qi, Zhaoyu Wang, Zhen Yang, Zhengxiao Du, Zhenyu Hou, and Zihan Wang. 2024.
\newblock \href {https://arxiv.org/abs/2406.12793} {Chatglm: A family of large language models from glm-130b to glm-4 all tools}.
\newblock \emph{Preprint}, arXiv:2406.12793.

\bibitem[{Hu et~al.(2021)Hu, Shen, Wallis, Allen-Zhu, Li, Wang, Wang, and Chen}]{hu2021lora}
Edward~J Hu, Yelong Shen, Phillip Wallis, Zeyuan Allen-Zhu, Yuanzhi Li, Shean Wang, Lu~Wang, and Weizhu Chen. 2021.
\newblock Lora: Low-rank adaptation of large language models.
\newblock \emph{arXiv preprint arXiv:2106.09685}.

\bibitem[{Hu et~al.(2024)Hu, Chen, and Ponti}]{hu2024fine}
Hanxu Hu, Pinzhen Chen, and Edoardo~M Ponti. 2024.
\newblock Fine-tuning large language models with sequential instructions.
\newblock \emph{arXiv preprint arXiv:2403.07794}.

\bibitem[{Jiang(2023)}]{jiang2023team}
Ye~Jiang. 2023.
\newblock Team qust at semeval-2023 task 3: A comprehensive study of monolingual and multilingual approaches for detecting online news genre, framing and persuasion techniques.
\newblock In \emph{Proceedings of the 17th International Workshop on Semantic Evaluation (SemEval-2023)}, pages 300--306.

\bibitem[{Jiang et~al.(2024)Jiang, Wang, Xu, Wang, Song, and Maynard}]{jiang2024cross}
Ye~Jiang, Taihang Wang, Xiaoman Xu, Yimin Wang, Xingyi Song, and Diana Maynard. 2024.
\newblock Cross-modal augmentation for few-shot multimodal fake news detection.
\newblock \emph{arXiv preprint arXiv:2407.12880}.

\bibitem[{Jiang and Wang(2024)}]{jiang2024large}
Ye~Jiang and Yimin Wang. 2024.
\newblock Large visual-language models are also good classifiers: A study of in-context multimodal fake news detection.
\newblock \emph{arXiv preprint arXiv:2407.12879}.

\bibitem[{Jiang et~al.(2023)Jiang, Yu, Wang, Xu, Song, and Maynard}]{jiang2023similarity}
Ye~Jiang, Xiaomin Yu, Yimin Wang, Xiaoman Xu, Xingyi Song, and Diana Maynard. 2023.
\newblock Similarity-aware multimodal prompt learning for fake news detection.
\newblock \emph{Information Sciences}, 647:119446.

\bibitem[{Jin et~al.(2022)Jin, Preo{\c{t}}iuc-Pietro, Do{\u{g}}ru{\"o}z, and Aletras}]{jin2022automatic}
Mali Jin, Daniel Preo{\c{t}}iuc-Pietro, A~Seza Do{\u{g}}ru{\"o}z, and Nikolaos Aletras. 2022.
\newblock Automatic identification and classification of bragging in social media.
\newblock In \emph{Proceedings of the 60th Annual Meeting of the Association for Computational Linguistics (Volume 1: Long Papers)}, pages 3945--3959.

\bibitem[{Juan Jos{\'e}~Bucher and Martini(2024)}]{juan2024fine}
Martin Juan Jos{\'e}~Bucher and Marco Martini. 2024.
\newblock Fine-tuned'small'llms (still) significantly outperform zero-shot generative ai models in text classification.
\newblock \emph{arXiv e-prints}, pages arXiv--2406.

\bibitem[{Meta(2024)}]{meta2024introducing}
AI~Meta. 2024.
\newblock Introducing meta llama 3: The most capable openly available llm to date, 2024.
\newblock \emph{URL https://ai. meta. com/blog/meta-llama-3/. Accessed on April}, 26.

\bibitem[{M{\o}ller and Aiello(2024)}]{moller2024prompt}
Anders~Giovanni M{\o}ller and Luca~Maria Aiello. 2024.
\newblock Prompt refinement or fine-tuning? best practices for using llms in computational social science tasks.
\newblock \emph{arXiv preprint arXiv:2408.01346}.

\bibitem[{Mosbach et~al.(2023)Mosbach, Pimentel, Ravfogel, Klakow, and Elazar}]{mosbach2023few}
Marius Mosbach, Tiago Pimentel, Shauli Ravfogel, Dietrich Klakow, and Yanai Elazar. 2023.
\newblock Few-shot fine-tuning vs. in-context learning: A fair comparison and evaluation.
\newblock In \emph{The 61st Annual Meeting Of The Association For Computational Linguistics}.

\bibitem[{Mu et~al.(2024)Mu, Wu, Thorne, Robinson, Aletras, Scarton, Bontcheva, and Song}]{mu2024navigating}
Yida Mu, Ben~P Wu, William Thorne, Ambrose Robinson, Nikolaos Aletras, Carolina Scarton, Kalina Bontcheva, and Xingyi Song. 2024.
\newblock Navigating prompt complexity for zero-shot classification: A study of large language models in computational social science.
\newblock In \emph{Proceedings of the 2024 Joint International Conference on Computational Linguistics, Language Resources and Evaluation (LREC-COLING 2024)}, pages 12074--12086.

\bibitem[{Naveed et~al.(2023)Naveed, Khan, Qiu, Saqib, Anwar, Usman, Akhtar, Barnes, and Mian}]{naveed2023comprehensive}
Humza Naveed, Asad~Ullah Khan, Shi Qiu, Muhammad Saqib, Saeed Anwar, Muhammad Usman, Naveed Akhtar, Nick Barnes, and Ajmal Mian. 2023.
\newblock A comprehensive overview of large language models.
\newblock \emph{arXiv preprint arXiv:2307.06435}.

\bibitem[{Ouyang et~al.(2022)Ouyang, Wu, Jiang, Almeida, Wainwright, Mishkin, Zhang, Agarwal, Slama, Ray et~al.}]{ouyang2022training}
Long Ouyang, Jeffrey Wu, Xu~Jiang, Diogo Almeida, Carroll Wainwright, Pamela Mishkin, Chong Zhang, Sandhini Agarwal, Katarina Slama, Alex Ray, et~al. 2022.
\newblock Training language models to follow instructions with human feedback.
\newblock \emph{Advances in neural information processing systems}, 35:27730--27744.

\bibitem[{Parthasarathy et~al.(2024)Parthasarathy, Zafar, Khan, and Shahid}]{parthasarathy2024ultimate}
Venkatesh~Balavadhani Parthasarathy, Ahtsham Zafar, Aafaq Khan, and Arsalan Shahid. 2024.
\newblock The ultimate guide to fine-tuning llms from basics to breakthroughs: An exhaustive review of technologies, research, best practices, applied research challenges and opportunities.
\newblock \emph{arXiv preprint arXiv:2408.13296}.

\bibitem[{Preo{\c{t}}iuc-Pietro et~al.(2019)Preo{\c{t}}iuc-Pietro, Gaman, and Aletras}]{preoctiuc2019automatically}
Daniel Preo{\c{t}}iuc-Pietro, Mihaela Gaman, and Nikolaos Aletras. 2019.
\newblock Automatically identifying complaints in social media.
\newblock In \emph{Proceedings of the 57th Annual Meeting of the Association for Computational Linguistics}, pages 5008--5019.

\bibitem[{Qin et~al.(2024)Qin, Yang, Guo, Li, Shao, Shi, Xu, Gu, Li, and Sun}]{qin2024unleashing}
Yulei Qin, Yuncheng Yang, Pengcheng Guo, Gang Li, Hang Shao, Yuchen Shi, Zihan Xu, Yun Gu, Ke~Li, and Xing Sun. 2024.
\newblock Unleashing the power of data tsunami: A comprehensive survey on data assessment and selection for instruction tuning of language models.
\newblock \emph{arXiv preprint arXiv:2408.02085}.

\bibitem[{Roy et~al.(2023)Roy, Harshvardhan, Mukherjee, and Saha}]{roy2023probing}
Sarthak Roy, Ashish Harshvardhan, Animesh Mukherjee, and Punyajoy Saha. 2023.
\newblock Probing llms for hate speech detection: strengths and vulnerabilities.
\newblock In \emph{Findings of the Association for Computational Linguistics: EMNLP 2023}, pages 6116--6128.

\bibitem[{Shu et~al.(2020)Shu, Mahudeswaran, Wang, Lee, and Liu}]{shu2020fakenewsnet}
Kai Shu, Deepak Mahudeswaran, Suhang Wang, Dongwon Lee, and Huan Liu. 2020.
\newblock Fakenewsnet: A data repository with news content, social context, and spatiotemporal information for studying fake news on social media.
\newblock \emph{Big data}, 8(3):171--188.

\bibitem[{Team et~al.(2024)Team, Riviere, Pathak, Sessa, Hardin, Bhupatiraju, Hussenot, Mesnard, Shahriari, Ram{\'e} et~al.}]{team2024gemma}
Gemma Team, Morgane Riviere, Shreya Pathak, Pier~Giuseppe Sessa, Cassidy Hardin, Surya Bhupatiraju, L{\'e}onard Hussenot, Thomas Mesnard, Bobak Shahriari, Alexandre Ram{\'e}, et~al. 2024.
\newblock Gemma 2: Improving open language models at a practical size.
\newblock \emph{arXiv preprint arXiv:2408.00118}.

\bibitem[{Wang et~al.(2024{\natexlab{a}})Wang, Ma, Feng, Zhang, Yang, Zhang, Chen, Tang, Chen, Lin et~al.}]{wang2024survey}
Lei Wang, Chen Ma, Xueyang Feng, Zeyu Zhang, Hao Yang, Jingsen Zhang, Zhiyuan Chen, Jiakai Tang, Xu~Chen, Yankai Lin, et~al. 2024{\natexlab{a}}.
\newblock A survey on large language model based autonomous agents.
\newblock \emph{Frontiers of Computer Science}, 18(6):186345.

\bibitem[{Wang et~al.(2024{\natexlab{b}})Wang, Yang, and Wei}]{wang2024learning}
Liang Wang, Nan Yang, and Furu Wei. 2024{\natexlab{b}}.
\newblock Learning to retrieve in-context examples for large language models.
\newblock In \emph{Proceedings of the 18th Conference of the European Chapter of the Association for Computational Linguistics (Volume 1: Long Papers)}, pages 1752--1767.

\bibitem[{Xu et~al.(2024)Xu, Li, Wang, Tian, and Jiang}]{xu-etal-2024-team}
Xiaoman Xu, Xiangrun Li, Taihang Wang, Jianxiang Tian, and Ye~Jiang. 2024.
\newblock \href {https://doi.org/10.18653/v1/2024.semeval-1.71} {Team {QUST} at {S}em{E}val-2024 task 8: A comprehensive study of monolingual and multilingual approaches for detecting {AI}-generated text}.
\newblock In \emph{Proceedings of the 18th International Workshop on Semantic Evaluation (SemEval-2024)}, pages 463--470, Mexico City, Mexico. Association for Computational Linguistics.

\bibitem[{Yang et~al.(2023{\natexlab{a}})Yang, Xiao, Wang, Zhang, Bian, Yin, Lv, Pan, Wang, Yan et~al.}]{yang2023baichuan}
Aiyuan Yang, Bin Xiao, Bingning Wang, Borong Zhang, Ce~Bian, Chao Yin, Chenxu Lv, Da~Pan, Dian Wang, Dong Yan, et~al. 2023{\natexlab{a}}.
\newblock Baichuan 2: Open large-scale language models.
\newblock \emph{arXiv preprint arXiv:2309.10305}.

\bibitem[{Yang et~al.(2024{\natexlab{a}})Yang, Yang, Hui, Zheng, Yu, Zhou, Li, Li, Liu, Huang et~al.}]{yang2024qwen2}
An~Yang, Baosong Yang, Binyuan Hui, Bo~Zheng, Bowen Yu, Chang Zhou, Chengpeng Li, Chengyuan Li, Dayiheng Liu, Fei Huang, et~al. 2024{\natexlab{a}}.
\newblock Qwen2 technical report.
\newblock \emph{arXiv preprint arXiv:2407.10671}.

\bibitem[{Yang et~al.(2024{\natexlab{b}})Yang, Gao, Ma, Lin, and Wang}]{yang2024reinforcement}
Ruichao Yang, Wei Gao, Jing Ma, Hongzhan Lin, and Bo~Wang. 2024{\natexlab{b}}.
\newblock Reinforcement tuning for detecting stances and debunking rumors jointly with large language models.
\newblock \emph{arXiv preprint arXiv:2406.02143}.

\bibitem[{Yang et~al.(2023{\natexlab{b}})Yang, Dai, Wang, and Sui}]{yang2023not}
Zhe Yang, Damai Dai, Peiyi Wang, and Zhifang Sui. 2023{\natexlab{b}}.
\newblock Not all demonstration examples are equally beneficial: Reweighting demonstration examples for in-context learning.
\newblock In \emph{Findings of the Association for Computational Linguistics: EMNLP 2023}, pages 13209--13221.

\bibitem[{Zhang et~al.(2023)Zhang, Dong, Li, Zhang, Sun, Wang, Li, Hu, Zhang, Wu et~al.}]{zhang2023instruction}
Shengyu Zhang, Linfeng Dong, Xiaoya Li, Sen Zhang, Xiaofei Sun, Shuhe Wang, Jiwei Li, Runyi Hu, Tianwei Zhang, Fei Wu, et~al. 2023.
\newblock Instruction tuning for large language models: A survey.
\newblock \emph{arXiv preprint arXiv:2308.10792}.

\bibitem[{Ziems et~al.(2024)Ziems, Held, Shaikh, Chen, Zhang, and Yang}]{ziems2024can}
Caleb Ziems, William Held, Omar Shaikh, Jiaao Chen, Zhehao Zhang, and Diyi Yang. 2024.
\newblock Can large language models transform computational social science?
\newblock \emph{Computational Linguistics}, 50(1):237--291.

\end{thebibliography}

\newpage

\appendix
\setcounter{table}{0}
\setcounter{figure}{0}
\renewcommand{\thefigure}{A\arabic{figure}}

\renewcommand{\thetable}{A\arabic{table}}
\onecolumn

\section{Appendix}

\subsection{Detailed performance comparison between LLMs on CSS tasks}

\label{sec:perf_vs}
\begin{figure*}[htbp]
\centering
\includegraphics[scale=0.164]{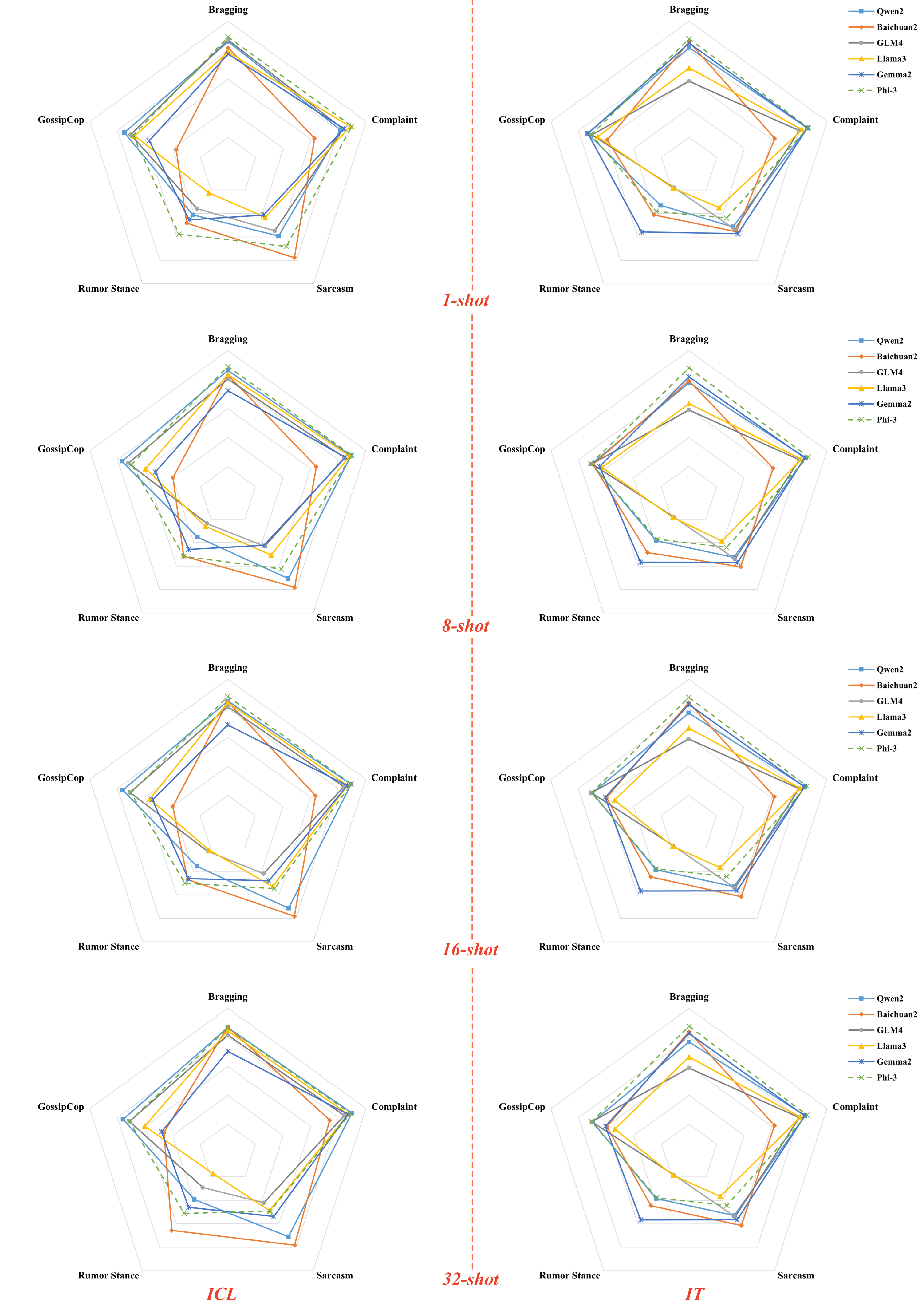} 
\caption{Performance comparison between LLMs on CSS tasks.}
\label{fig:model_vs}
\end{figure*}

\newpage
\subsection{Examples of instruction and constraints in CSS tasks}
\label{sec:ins&con}


\begin{table*}[htbp]
\large
\resizebox{\textwidth}{!}{%
\begin{tabular}{|l|p{7cm}|p{8cm}|}
\hline
\textbf{Dataset} & \textbf{Instruction}                                                                                                                                                                                            & \textbf{Constraint}                                                                                         \\ \hline
Bragging         & Analyze the content and determine if it includes a bragging statement.                                                                                                                                          & Respond only with bragging or not bragging , without providing any additional context or explanation.       \\ \hline
Complaint        & Analyze the content and determine if it includes a complaint.                                                                                                                                                   & Respond only with complaint or not complaint, without providing any additional context or explanation.      \\ \hline
Sarcasm          & Analyze the content and determine if it includes sarcasm.                                                                                                                                                       & Respond only with sarcasm or not sarcasm, without providing any additional context or explanation.          \\ \hline
Rumour Stance    & Given a tweet related to a rumour, classify its stance as one of the following categories: support, deny, query, or comment. Each tweet should only be associated with one stance category based on its content. & Respond only with the appropriate stance category, without providing any additional context or explanation. \\ \hline
GossipCop        & Given a news article, classify its truthfulness as either real or fake.                                                                                                                                         & Respond only with real or fake, without providing any additional context or explanation.                    \\ \hline
\end{tabular}%
}
\caption{Detailed information on the instructions and constraints for each benchmark task.}
\label{tab:ins&con}
\end{table*}

\subsection{Examples of CoT description}
\label{sec:cot}


\begin{table*}[htbp]
\footnotesize
\begin{tabular}{|p{3cm}|p{12.1cm}|}
\hline
\textbf{Tweet}                                                                                       & \textbf{CoT Strategy}                                                                                                                                                             \\ \hline
For a minute I was tired of being the bigger man , until I realized that 's just who I am . & Label: Bragging. Let’s break down the analysis. The tweet states, 'For a minute I was tired of being the bigger man, until I realized that's just who I am.' This sentence starts by highlighting a moment of frustration about always being 'the bigger man,' a common expression for acting maturely or taking the high road in difficult situations. The statement concludes by reaffirming the tweeter's inherent nature of being superior or more mature, which implies a self-promotion of personal virtues. The aim is to assess if this self-reflection serves to subtly brag about one’s character. Examining the structure and content, the tweet pivots to emphasize personal qualities, suggesting that the individual views these traits as defining and superior. The conclusion, based on the content's focus on self-praise, categorizes this tweet as bragging. \\ 
\hline
If the market is so efficient , why are n't we green yet ?                                                         & Label: Not Bragging. Analyzing the tweet: 'If the market is so efficient, why aren't we green yet?' This tweet poses a critical question about the efficiency of markets in achieving environmental sustainability. It is rhetorical and serves to challenge or provoke thought rather than to boast about personal achievements or qualities. The focus is on questioning societal or economic systems, highlighting a discrepancy between claimed efficiency and environmental outcomes. The purpose of this tweet is to engage others in reflection or debate regarding the effectiveness of market mechanisms in driving sustainable practices, not to showcase any personal accomplishment.                                                                                                                                                                                                                            \\ \hline
\end{tabular}%

\caption{The CoT strategy for Bragging samples.}
\label{tab:ad2}
\end{table*}


\subsection{Detailed experimental results}
\label{sec:results}


\renewcommand{\arraystretch}{2.5}
\definecolor{myblue}{rgb}{0.8, 0.85, 0.95}

\begin{table*}[htbp]
\small
\resizebox{\textwidth}{!}{
\setlength{\tabcolsep}{1.3mm}{
\begin{tabular}{ccccccccccc}
\hline
\multicolumn{11}{c}{\textit{\textbf{Qwen2}}}                                                                                                                                                                                                                                                      \\ \hline
\multirow{2}{*}{\textbf{\begin{tabular}[c]{@{}c@{}}n-shot\\    (seed)\end{tabular}}} & \multicolumn{2}{c}{\textbf{Bragging}} & \multicolumn{2}{c}{\textbf{Complaint}} & \multicolumn{2}{c}{\textbf{Sarcasm}} & \multicolumn{2}{c}{\textbf{Rumour Stance}} & \multicolumn{2}{c}{\textbf{GossipCop}} \\ \cline{2-11} 
                                                                                     & \textbf{ICL}       & \textbf{IT}      & \textbf{ICL}       & \textbf{IT}       & \textbf{ICL}      & \textbf{IT}      & \textbf{ICL}         & \textbf{IT}        & \textbf{ICL}       & \textbf{IT}       \\ \hline
\textbf{1(42)}                                                                       & 88.7/88.8          & 86.5/87.4        & 82.9/83.3          & 85.7/86.0           & 64.0/67.1           & 56.9/60.3        & 44.3/46.7            & 37.3/30.6          & 75.7/63.5          & 73.3/60.8         \\
\textbf{1(43)}                                                                       & 84.0/85.6            & 77.2/80.9        & 77.4/77.8          & 85.2/85.5         & 56.8/60.2         & 50.5/53.4        & 36.2/39.3            & 36.4/29.1          & 74.1/58.1          & 73.3/60.8         \\
\textbf{1(44)}                                                                       & 87.2/87.5          & 86.3/87.3        & 83.6/84.0            & 83.8/84.1         & 64.9/67.9         & 50.9/53.8        & 42.6/44.8            & 28.5/25.3          & 75.2/60.0            & 73.3/60.8         \\
\textbf{1(45)}                                                                       & 85.2/86.5          & 82.0/84.3          & 86.7/86.9          & 89.0/89.2           & 52.5/56.0           & 42.9/43.9        & 46.9/50.2            & 36.8/29.6          & 76.5/63.4          & 74.0/54.1           \\
\textbf{1(46)}                                                                       & 84.7/86.1          & 76.5/80.3        & 76.8/77.2          & 84.5/84.8         & 58.2/61.5         & 55.9/59.1        & 36.4/39.2            & 26.9/24.8          & 73.7/58.0            & 73.6/58.9         \\
\textbf{Avg}                                                                         & 86.0/86.9            & 81.7/84.0          & 81.5/81.8          & 85.6/85.9         & 59.3/62.5         & 51.4/54.1        & 41.3/44.0              & 33.2/27.9          & 75.0/60.6            & 73.5/59.1         \\ \hline
\textbf{8(42)}                                                                       & 86.6/87.5          & 76.2/80.1        & 89.6/89.7          & 84.1/84.4         & 74.5/74.6         & 51.8/55.1        & 30.3/33.1            & 40.3/30.8          & 76.0/62.2            & 66.2/60.4         \\
\textbf{8(43)}                                                                       & 86.1/87.3          & 71.4/76.4        & 89.6/89.7          & 85.1/85.4         & 71.1/72.8         & 51.3/54.3        & 40.4/44.4            & 36.5/30.1          & 75.9/62.6          & 68.5/61.6         \\
\textbf{8(44)}                                                                       & 86.7/87.5          & 78.4/81.7        & 89.3/89.4          & 85.8/86.1         & 70.6/72.6         & 52.6/55.7        & 36.5/40.6            & 35.1/28.6          & 76.9/64.5          & 70.7/62.9         \\
\textbf{8(45)}                                                                       & 85.9/87.0            & 82.1/84.3        & 89.9/89.9          & 85.1/85.4         & 69.5/71.8         & 55.4/58.9        & 32.6/35.9            & 37.2/30.4          & 76.7/62.0            & 74.2/61.2         \\
\textbf{8(46)}                                                                       & 86.2/87.2          & 80.3/83.1        & 87.5/87.8          & 82.9/83.3         & 68.2/71.0           & 52.4/55.3        & 37.7/41.7            & 43.1/32.9          & 77.2/64.1          & 72.4/62.3         \\
\textbf{Avg}                                                                         & 86.3/87.3          & 77.7/81.1        & 89.2/89.3          & 84.6/84.9         & 70.8/72.6         & 52.7/55.8        & 35.5/39.1            & 38.4/30.6          & 76.5/63.1          & 70.4/61.7         \\\hline
\textbf{16(42)}                                                                      & 86.0/87.0              & 79.4/82.5        & 91.3/91.4          & 83.2/83.6         & 70.4/72.3         & 54.3/57.6        & 33.1/36.6            & 37.1/29.5          & 76.1/61.8          & 71.3/62.4         \\
\textbf{16(43)}                                                                      & 85.1/86.6          & 71.3/76.3        & 88.1/88.3          & 84.6/85.0           & 71.5/73.2         & 55.9/59.2        & 37.5/41.6            & 39.4/31.3          & 76.6/64.1          & 67.3/61.0           \\
\textbf{16(44)}                                                                      & 82.9/84.9          & 77.9/81.4        & 88.6/88.8          & 84.8/85.1         & 71.5/73.2         & 52.5/55.5        & 34.6/38.5            & 36.5/29.5          & 76.1/62.8          & 67.4/61.2         \\
\textbf{16(45)}                                                                      & 83.1/85.2          & 75.2/79.3        & 90.1/90.3          & 83.5/83.8         & 68.4/70.9         & 49.9/52.6        & 31.1/34.1            & 38.2/30.5          & 75.5/60.0            & 72.7/62.1         \\
\textbf{16(46)}                                                                      & 87.5/88.1          & 80.0/82.9          & 88.8/89.0            & 84.3/84.7           & 75.7/75.8         & 52.3/55.4        & 42.9/47.2            & 41.7/32.2          & 75.9/60.6          & 70.2/62.4         \\
\textbf{Avg}                                                                         & 84.9/86.4          & 76.8/80.5        & 89.4/89.5          & 84.1/84.4         & 71.5/73.1         & 53.0/56.1          & 35.8/39.6            & 38.6/30.6          & 76.1/61.9            & 69.8/61.8         \\\hline
\textbf{32(42)}                                                                      & 87.2/88.0            & 76.3/80.2        & 89.9/89.9          & 84.6/85.0           & 69.9/71.6         & 52.2/55.3        & 40.2/44.4            & 47.1/37.4          & 75.4/62.4          & 67.3/61.3         \\
\textbf{32(43)}                                                                      & 86.0/87.3            & 73.9/78.3        & 90.1/90.3          & 84.1/84.4         & 70.3/72.3         & 52.7/55.7        & 39.6/43.4            & 46.0/35.5            & 76.3/62.4          & 69.7/62.2         \\
\textbf{32(44)}                                                                      & 86.2/87.3          & 78.4/81.8        & 89.9/90.0            & 83.2/83.6         & 72.6/74.0           & 53.6/56.9        & 42.5/46.8            & 48.7/36.3          & 75.7/61.0            & 71.3/61.7         \\
\textbf{32(45)}                                                                      & 86.7/87.7          & 75.9/79.9        & 90.1/90.3          & 85.9/86.2         & 69.5/71.5         & 51.5/54.5        & 37.4/41.7            & 46.3/36.6          & 76.4/63.3          & 73.3/59.9         \\
\textbf{32(46)}                                                                      & 87.3/88.1          & 76.6/80.4        & 89.6/89.7          & 82.9/83.3         & 73.0/74.1           & 48.6/51.2        & 37.0/40.8              & 44.9/35.7          & 75.3/61.3          & 71.4/62.9         \\
\textbf{Avg}                                                                         & 86.7/87.7          & 76.2/80.1        & 89.9/90.0            & 84.1/84.5         & 71.1/72.7         & 51.7/54.7        & 39.4/43.4            & 46.6/36.3          & 75.8/62.1          & 70.6/61.6 \\ \hline      
\end{tabular}}
}
\caption{The detailed experimental results for Qwen2.}
\label{tab:re7}
\end{table*}

\renewcommand{\arraystretch}{2.5}
\definecolor{myblue}{rgb}{0.8, 0.85, 0.95}

\begin{table*}[]
\small
\setlength{\tabcolsep}{1.3mm}{
\begin{tabular}{ccccccccccc}
\hline
\multicolumn{11}{c}{\textit{\textbf{Baichuan2}}}                                                                                                                                                                                                                                                                                                                                         \\ \hline
\multirow{2}{*}{\textbf{\begin{tabular}[c]{@{}c@{}}n-shot\\    (seed)\end{tabular}}} & \multicolumn{2}{c}{\textbf{Bragging}} & \multicolumn{2}{c}{\textbf{Complaint}} & \multicolumn{2}{c}{\textbf{Sarcasm}} & \multicolumn{2}{c}{\textbf{Rumour Stance}} & \multicolumn{2}{c}{\textbf{GossipCop}} \\ \cline{2-11} 
                                                                                     & \textbf{ICL}       & \textbf{IT}      & \textbf{ICL}       & \textbf{IT}       & \textbf{ICL}      & \textbf{IT}      & \textbf{ICL}         & \textbf{IT}        & \textbf{ICL}       & \textbf{IT}       \\ \hline
\textbf{1(42)}                                                                         & 83.9/57.8          & 87.9/56.7        & 71.8/69.0            & 69.6/66.8         & 78.1/43.9         & 62.4/45.6        & 52.4/37.6            & 39.9/28.3          & 34.5/33.3          & 54.3/51.0           \\
\textbf{1(43)}                                                                         & 87.3/64.1          & 88.0/54.2          & 71.0/70.7            & 37.4/30.0           & 78.1/45.2         & 43.3/40.7        & 44.3/29.8            & 42.8/31.4          & 50.6/50.0            & 54.3/51.0           \\
\textbf{1(44)}                                                                         & 83.2/59.8          & 88.3/56.7        & 47.7/46.2          & 70.1/67.0           & 76.8/52.9         & 69.8/53.6        & 51.5/32.7            & 41.2/29.8          & 34.8/33.8          & 54.3/51.0           \\
\textbf{1(45)}                                                                         & 88.4/67.3          & 85.6/64.7        & 51.7/51.0            & 69.3/63.5         & 77.8/55.7         & 73.1/47.2        & 42.5/29.4            & 40.5/28.1          & 32.6/30.7          & 72.5/42.6         \\
\textbf{1(46)}                                                                         & 65.7/55.1          & 78.9/61.5        & 70.7/66.7          & 63.7/63.3         & 77.6/56.6         & 28.5/27.8        & 51.2/34.5            & 41.5/31.4          & 35.4/34.3          & 60.8/54.4         \\
\textbf{Avg}                                                                           & 81.7/60.8          & 85.7/58.8        & 62.6/60.7          & 62.0/58.1            & 77.7/50.9         & 55.4/43.0          & 48.4/32.8            & 41.2/29.8          & 37.6/36.4          & 59.2/50.0           \\ \hline
\textbf{8(42)}                                                                         & 80.0/62.2            & 87.8/63.7        & 68.3/66.6          & 62.6/59.7         & 78.1/43.9         & 68.9/56.4         & 59.9/29.7            & 50.7/30.5          & 41.2/41.2          & 66.6/53.4         \\
\textbf{8(43)}                                                                         & 79.2/65.1          & 73.9/57.9        & 74.3/73.7          & 60.7/57.4         & 78.2/44.4         & 56.2/51.7        & 57.9/34.1            & 41.3/29.5          & 40.0/40.0              & 67.8/54.3         \\
\textbf{8(44)}                                                                         & 86.8/65.7          & 78.9/59.2        & 60.0/56.8            & 62.5/51.1         & 78.2/44.4         & 55.7/51.9        & 47.5/28.1            & 62.9/24.9          & 39.3/39.2          & 68.5/53.6         \\
\textbf{8(45)}                                                                         & 83.0/65.8            & 72.7/55.2        & 38.0/31.1            & 57.8/53.1         & 78.2/44.4         & 65.1/54.3        & 53.1/29.8            & 35.4/27.0            & 39.7/39.6          & 72.0/47.6           \\
\textbf{8(46)}                                                                         & 86.8/67.1          & 81.9/61.7        & 79.9/79.1          & 61.5/44.5         & 78.2/44.3         & 57.7/50.2        & 40.2/25.8            & 52.7/30.1          & 38.5/38.2          & 72.1/52.6         \\
\textbf{Avg}                                                                           & 83.1/65.2          & 79.0/59.6          & 64.1/61.5          & 61.0/53.2            & 78.2/44.3         & 60.7/52.9        & 51.7/29.5            & 48.6/28.4          & 39.8/39.6          & 69.4/52.3         \\ \hline
\textbf{16(42)}                                                                        & 77.4/62.2          & 85.9/63.9        & 50.1/50.0            & 62.3/57.1         & 78.2/44.4         & 62.954.7         & 52.9/29.0              & 41.4/29.3          & 47.2/46.5          & 57.8/52.1         \\
\textbf{16(43)}                                                                        & 84.8/69.8          & 83.4/61.1        & 64.3/55.6          & 60.1/50.0           & 78.1/43.9         & 60.8/53.1        & 50.1/29.9            & 43.8/28.2          & 38.9/38.8          & 68.9/55.9         \\
\textbf{16(44)}                                                                        & 90.8/70.6          & 81.4/61.9        & 71.1/67.0            & 62.7/54.3         & 78.2/44.4         & 60.7/51.3        & 46.1/24.9            & 48.3/31.6          & 38.0/37.8            & 58.2/52.8         \\
\textbf{16(45)}                                                                        & 83.7/66.9          & 83.0/63.2          & 64.1/59.1          & 62.4/54.8         & 78.1/43.9         & 59.2/50.3        & 46.9/26.9            & 40.1/30.8          & 40.6/40.6          & 72.6/42.6         \\
\textbf{16(46)}                                                                        & 85.7/67.4          & 85.4/63.6        & 68.6/55.0            & 62.3/51.5         & 78.2/44.4         & 63.8/53.8        & 38.8/23.7            & 49.4/28.2          & 34.4/33.6          & 40.5/40.4         \\
\textbf{Avg}                                                                           & 84.5/67.4          & 83.8/62.7        & 63.6/57.3          & 62.0/53.5            & 78.2/44.2         & 61.5/52.6        & 47.0/26.9            & 44.6/29.6          & 39.8/39.5          & 59.6/48.8        \\ \hline
\textbf{32(42)}                                                                        & 88.8/66.5          & 83.8/63.1        & 72.2/64.4          & 61.8/51.2         & 78.0/44.3           & 62.9/56.2         & 65.7/27.4            & 52.9/29.6          & 42.6/42.5          & 36.8/36.2         \\
\textbf{32(43)}                                                                        & 87.2/67.0            & 79.5/59.8        & 71.6/65.0            & 65.3/58.7         & 78.2/44.4         & 58.1/52.9        & 66.2/25.1            & 38.6/29.1          & 51.0/48.7            & 67.7/54.6         \\
\textbf{32(44)}                                                                        & 86.0/68.5            & 78.5/58.8        & 68.2/56.1          & 63.5/56.3         & 78.2/44.4         & 64.4/55.5        & 65.2/25.7            & 46.1/29.9          & 43.3/43.0            & 72.3/48.2         \\
\textbf{32(45)}                                                                        & 87.1/67.4          & 83.9/62.4        & 78.1/75.2          & 60.4/52.0           & 78.2/44.4         & 60.8/52.3        & 64.6/23.8            & 34.1/28.5          & 45.0/44.6            & 72.7/44.5         \\
\textbf{32(46)}                                                                        & 86.6/67.8          & 78.0/59.3          & 78.7/78.1          & 61.2/49.5         & 78.1/43.9         & 58.8/52.6        & 66.7/28.0              & 50.1/30.6          & 49.2/48.3          & 40.5/40.5         \\
\textbf{Avg}                                                                           & 87.1/67.4          & 80.8/60.7        & 73.7/67.8          & 62.4/53.5         & 78.2/44.2         & 61.0/53.9          & 65.7/26.0              & 44.4/29.5          & 46.2/45.4          & 58.0/44.8  \\ \hline        
\end{tabular}}
\caption{The detailed experimental results for Baichuan2.}
\label{tab:re8}
\end{table*}
\renewcommand{\arraystretch}{2.5}
\definecolor{myblue}{rgb}{0.8, 0.85, 0.95}

\begin{table*}[]
\small
\setlength{\tabcolsep}{1.3mm}{
\begin{tabular}{ccccccccccc}
\hline
\multicolumn{11}{c}{\textit{\textbf{GLM4}}}                                                                                                                                                                                                                                                         \\ \hline
\multirow{2}{*}{\textbf{\begin{tabular}[c]{@{}c@{}}n-shot\\    (seed)\end{tabular}}} & \multicolumn{2}{c}{\textbf{Bragging}} & \multicolumn{2}{c}{\textbf{Complaint}} & \multicolumn{2}{c}{\textbf{Sarcasm}} & \multicolumn{2}{c}{\textbf{Rumour Stance}} & \multicolumn{2}{c}{\textbf{GossipCop}} \\ \cline{2-11} 
                                                                                     & \textbf{ICL}       & \textbf{IT}      & \textbf{ICL}       & \textbf{IT}       & \textbf{ICL}      & \textbf{IT}      & \textbf{ICL}         & \textbf{IT}        & \textbf{ICL}       & \textbf{IT}       \\ \hline
\textbf{1(42)}                                                                         & 84.7/71.6          & 58.8/51.8        & 83.0/82.4            & 79.7/79.3         & 56.5/55.0           & 53.9/51.6        & 35.4/27.8            & 17.9/16.1          & 72.2/61.5          & 70.3/57.8         \\
\textbf{1(43)}                                                                         & 86.5/72.6          & 58.6/51.7        & 85.7/84.8          & 79.7/79.3         & 49.2/48.7         & 53.3/51.1        & 41.9/30.6            & 17.8/15.9          & 68.2/60.4          & 70.3/57.8         \\
\textbf{1(44)}                                                                         & 90.5/75.6          & 58.6/51.8        & 82.9/82.5          & 80.1/79.7         & 54.0/52.8           & 53.3/51.1        & 37.9/29.6            & 17.8/15.9          & 69.3/63.0            & 70.3/57.8         \\
\textbf{1(45)}                                                                         & 85.9/73.0            & 58.6/51.7        & 83.6/83.2          & 79.9/79.6         & 51.1/50.5         & 53.9/51.6        & 31.1/25.8            & 17.9/16.1          & 73.3/61.7          & 70.3/57.8         \\
\textbf{1(46)}                                                                         & 87.7/75.3          & 58.6/51.8        & 81.7/81.5          & 79.7/79.3         & 63.2/60.1         & 53.3/51.1        & 34.3/27.6            & 17.9/16.1          & 69.5/61.9          & 70.5/57.8         \\
\textbf{Avg}                                                                           & 87.1/73.7            & 58.6/51.8        & 83.4/82.9          & 79.8/79.4         & 54.8/53.4         & 53.5/51.3        & 36.1/28.3            & 17.9/16.0            & 70.5/61.7          & 70.4/57.8         \\ \hline
\textbf{8(42)}                                                                         & 77.4/66.1          & 59.1/52.1         & 83.3/83.0            & 79.8/79.4          & 44.5/44.3         & 54.2/52.0           & 22.8/22.3            & 18.1/16.0            & 71.8/63.8          & 70.3/57.8         \\
\textbf{8(43)}                                                                         & 79.8/68.3          & 59.1/52.1        & 84.5/84.1          & 79.9/79.6         & 40.6/40.6         & 53.8/51.5        & 22.1/22.8            & 18.1/16.1          & 74.4/62.9          & 70.3/57.8         \\
\textbf{8(44)}                                                                         & 80.5/68.6          & 59.1/52.2        & 86.8/86.4          & 79.8/79.5         & 42.4/42.4         & 54.0/51.9          & 23.2/23 .0             & 18.0/16.1            & 71.1/64.8          & 70.4/57.9         \\
\textbf{8(45)}                                                                         & 81.1/68.7          & 58.8/51.9        & 85.9/85.5          & 80.2/79.9         & 45.2/45.0           & 54.2/52.0          & 26.4/27.0              & 18.0/16.1            & 71.4/63.1          & 70.3/57.8         \\
\textbf{8(46)}                                                                         & 82.3/70.1          & 59.1/52.1        & 81.4/81.2          & 79.8/79.4         & 43.5/43.4         & 54.1/51.9        & 25.5/24.7            & 17.8/15.9          & 71.8/63.8          & 70.2/57.7         \\
\textbf{Avg}                                                                           & 80.2/68.4          & 59.0/52.1          & 84.4/84.0            & 79.9/79.5         & 43.2/43.2         & 54.1/51.9        & 24.0/24.0                & 18.0/16.0              & 72.1/63.7          & 70.3/57.8         \\ \hline
\textbf{16(42)}                                                                        & 76.4/65.0            & 59.2/52.2         & 85.1/84.6          & 79.8/79.4          & 42.0/42.0             & 54.1/51.8         & 22.0/23.2              & 18.2/16.1          & 71.7/64.8          & 70.4/57.9         \\
\textbf{16(43)}                                                                        & 80.7/69.3          & 58.6/51.8        & 84.1/83.7          & 79.9/79.6         & 45.6/45.4         & 53.8/51.6        & 25.0/25.7              & 18.1/16.0            & 70.4/63.9          & 70.3/57.8         \\
\textbf{16(44)}                                                                        & 83.8/70.8          & 58.6/51.8        & 84.8/84.4          & 79.8/79.4         & 38.8/38.8         & 53.8/51.6        & 22.9/23.6            & 18.1/16.0            & 70.1/64.6          & 70.4/57.9         \\
\textbf{16(45)}                                                                        & 79.0/67.4            & 58.8/51.8        & 84.6/84.3          & 80.1/79.7         & 40.1/40.1         & 54.2/52.0          & 23.6/24.3            & 18.2/16.1          & 70.3/63.9          & 70.4/57.9         \\
\textbf{16(46)}                                                                        & 85.5/73.5          & 59.0/52.0            & 82.2/81.9          & 79.7/79.3         & 43.7/43.7         & 54.3/52.0          & 22.0/23.6              & 18.0/15.9            & 71.6/64.3          & 70.5/57.8         \\
\textbf{Avg}                                                                           & 81.1/69.2          & 58.8/51.9        & 84.1/83.8          & 79.8/79.5         & 42.1/42.0           & 54.1/51.8          & 23.1/24.1            & 18.1/16.0            & 70.8/64.3          & 70.4/57.9         \\ \hline
\textbf{32(42)}                                                                        & 78.4/66.7          & 58.251.5         & 85.5/85.1          & 79.9/79.6          & 45.4/45.2         & 54.1/51.7         & 27.8/28.0              & 17.7/15.7          & 72.2/64.4          & 70.1/57.6         \\
\textbf{32(43)}                                                                        & 84.1/72.0            & 58.8/51.9        & 85.4/85.0            & 79.9/79.6         & 41.0/41.0             & 53.7/51.5        & 30.9/30.0              & 17.7/15.7          & 72.1/65.8          & 70.4/57.8         \\
\textbf{32(44)}                                                                        & 80.5/68.6          & 58.9/51.9        & 83.2/82.9          & 79.7/79.3         & 44.3/44.2         & 54.2/52.0          & 28.2/25.6            & 17.9/16.1          & 70.6/64.6          & 70.5/57.9         \\
\textbf{32(45)}                                                                        & 80.5/68.6          & 58.8/51.9        & 84.2/83.9          & 79.8/79.4         & 38.2/38.1         & 54.0/51.8          & 25.0/25.7              & 18.2/16.3          & 70.0/64.0              & 70.5/57.8         \\
\textbf{32(46)}                                                                        & 83.5/70.9          & 59.1/52.2        & 83.5/83.2          & 79.8/79.4         & 41.8/41.8         & 53.7/51.4        & 34.3/30.8            & 18.1/16.3          & 71.0/64.6            & 70.4/58.0           \\
\textbf{Avg}                                                                           & 81.4/69.4          & 58.8/51.9        & 84.4/84.0            & 79.8/79.5         & 42.1/42.1           & 53.9/51.7        & 29.2/28.0              & 17.9/16.0            & 71.2/64.7          & 70.4/57.8  \\ \hline      
\end{tabular}}
\caption{The detailed experimental results for GLM4.}
\label{tab:re9}
\end{table*}

\renewcommand{\arraystretch}{2.5}
\definecolor{myblue}{rgb}{0.8, 0.85, 0.95}

\begin{table*}[]
\small
\setlength{\tabcolsep}{1.3mm}{
\begin{tabular}{ccccccccccc}
\hline
\multicolumn{11}{c}{\textit{\textbf{Llama3}}}                                                                                                                                                                                                                                                       \\ \hline
\multirow{2}{*}{\textbf{\begin{tabular}[c]{@{}c@{}}n-shot\\    (seed)\end{tabular}}} & \multicolumn{2}{c}{\textbf{Bragging}} & \multicolumn{2}{c}{\textbf{Complaint}} & \multicolumn{2}{c}{\textbf{Sarcasm}} & \multicolumn{2}{c}{\textbf{Rumour Stance}} & \multicolumn{2}{c}{\textbf{GossipCop}} \\ \cline{2-11} 
                                                                                     & \textbf{ICL}       & \textbf{IT}      & \textbf{ICL}       & \textbf{IT}       & \textbf{ICL}      & \textbf{IT}      & \textbf{ICL}         & \textbf{IT}        & \textbf{ICL}       & \textbf{IT}       \\ \hline
\textbf{1(42)}                                                                         & 81.4/66.7          & 78.4/62.9        & 88.7/88.1          & 81.7/81.2         & 44.6/44.5         & 36.4/36.1        & 15.0/14.7              & 19.0/19.7            & 70.0/59.2            & 65.0/58.6           \\
\textbf{1(43)}                                                                         & 80.6/65.7          & 59.8/51.9        & 88.0/87.2            & 80.8/80.4         & 46.1/46.0           & 37.3/37.0         & 23.7/19.9            & 16.5/17.2          & 72.2/59.4          & 65.0/58.6           \\
\textbf{1(44)}                                                                         & 78.1/63.4          & 64.9/55.3        & 89.8/89.1          & 82.0/81.5           & 48.5/48.2         & 43.0/43.0            & 35.6/25.6            & 18.4/20.0            & 70.9/58.3          & 65.0/58.6           \\
\textbf{1(45)}                                                                         & 74.9/62.4          & 70.6/58.4        & 88.7/88.1          & 84.5/83.3         & 37.5/37.3         & 28.0/26.2          & 17.7/16.4            & 19.4/21.2          & 66.8/62.0            & 71.7/56.1         \\
\textbf{1(46)}                                                                         & 78.7/64.6          & 64.7/55.0          & 86.9/86.4          & 80.1/79.8         & 40.1/40.1         & 30.5/29.3        & 20.3/16.2            & 19.4/22.7          & 58.0/55.0              & 64.5/57.5         \\
\textbf{Avg}                                                                           & 78.7/64.6          & 67.7/56.7        & 88.4/87.8          & 81.8/81.3         & 43.4/43.2         & 35.1/34.3        & 22.5/18.6            & 18.6/20.2          & 67.6/58.8          & 66.2/57.9         \\ \hline
\textbf{8(42)}                                                                         & 82.6/68.1          & 66.1/55.7        & 86.1/83.9          & 81.9/81.3         & 55.4/54.2         & 34.9/34.4        & 26.0/19.6              & 18.5/19.1          & 62.6/59.0            & 69.2/58.3         \\
\textbf{8(43)}                                                                         & 81.6/67.4          & 57.5/50.5        & 88.2/86.8          & 80.3/79.9         & 48.0/47.9           & 30.5/29.3        & 16.8/15.3            & 18.2/19.7          & 64.5/58.9          & 65.8/58.3         \\
\textbf{8(44)}                                                                         & 82.6/66.6          & 68.8/57.4        & 88.7/87.5          & 81.9/81.3         & 51.8/51.3         & 47.7/47.6        & 32.5/23.9            & 18.5/20.1          & 55.5/53.6          & 51.7/50.7         \\
\textbf{8(45)}                                                                         & 85.2/70.6          & 56.9/49.9        & 87.1/85.2          & 82.7/82.0           & 54.6/53.7         & 40.0/40.0            & 31.3/23.7            & 18.8/20.3          & 62.5/58.8          & 64.3/58.3         \\
\textbf{8(46)}                                                                         & 84.8/70.6          & 67.5/56.5        & 90.4/89.5          & 79.8/79.5         & 43.9/43.9         & 40.6/40.5        & 25.0/19.5              & 19.3/19.7          & 53.0/51.6            & 63.9/58.3         \\
\textbf{Avg}                                                                           & 83.3/68.6          & 63.4/54.0          & 88.1/86.6          & 81.3/80.8         & 50.8/50.2         & 38.8/38.4        & 26.3/20.4            & 18.7/19.8          & 59.6/56.4          & 63.0/56.8           \\ \hline
\textbf{16(42)}                                                                        & 83.6/67.9          & 64.3/54.8        & 85.1/81.9          & 79.7/79.4         & 59.5/57.7         & 39.4/39.4        & 15.9/13.6            & 18.8/19.9          & 58.6/56.2          & 61.5/57.0           \\
\textbf{16(43)}                                                                        & 80.7/67.1          & 65.3/55.4        & 86.9/85.0            & 81.0/80.5           & 49.3/49.0           & 33.2/32.5        & 20.1/17.3            & 19.0/19.2            & 56.1/54.4          & 60.1/56.5         \\
\textbf{16(44)}                                                                        & 83.7/67.5          & 71.5/59.0          & 86.9/85.0            & 81.9/81.4         & 53.5/52.7         & 37.5/37.3        & 26.2/19.9            & 18.1/19.0            & 52.7/51.6          & 41.1/41.1         \\
\textbf{16(45)}                                                                        & 85.9/71.6          & 59.5/51.6        & 87.1/85.0            & 80.8/80.4         & 54.6/53.7         & 35.2/34.8        & 27.7/21.2            & 18.1/18.5          & 60.5/57.4          & 60.8/56.6         \\
\textbf{16(46)}                                                                        & 85.6/70.0            & 71.2/58.9        & 87.1/84.9          & 80.1/79.8         & 47.3/47.2         & 37.8/37.6        & 19.8/16.9            & 19.0/19.4            & 54.5/53.1          & 45.3/45.3         \\
\textbf{Avg}                                                                           & 83.9/68.8          & 66.4/55.9        & 86.6/84.4          & 80.7/80.3         & 52.9/52.1         & 36.6/36.3        & 21.9/17.8            & 18.6/19.2          & 56.5/54.5          & 53.8/51.3          \\ \hline
\textbf{32(42)}                                                                        & 82.0/67.1            & 66.0/55.9          & 87.7/85.7          & 82.3/81.7         & 50.1/49.7         & 34.4/33.9        & 17.4/14.7            & 27.1/23.2          & 63.7/58.8          & 58.4/55.1         \\
\textbf{32(43)}                                                                        & 86.2/71.5          & 63.4/54.2        & 87.2/85.1          & 81.0/80.5           & 47.0/46.9           & 35.0/34.6          & 19.7/16.2            & 21.5/20.5          & 56.1/54.3          & 59.0/56.0             \\
\textbf{32(44)}                                                                        & 85.2/68.0            & 66.3/55.8        & 86.9/84.9          & 81.1/80.7         & 47.0/46.9           & 38.4/38.3        & 16.7/14.7            & 26.0/23.8            & 64.2/59.5          & 39.1/38.8         \\
\textbf{32(45)}                                                                        & 85.8/71.9          & 63.6/54.3        & 85.9/83.6          & 81.9/81.3         & 52.6/52.0           & 36.2/35.9        & 15.2/14.2            & 26.8/24.7          & 57.7/55.2          & 64.1/58.3         \\
\textbf{32(46)}                                                                        & 83.1/68.7          & 63.1/54.0          & 87.8/86.2          & 81.7/81.2         & 47.5/47.4         & 37.8/37.6        & 17.4/16.1            & 29.1/26.8          & 59.8/57.0            & 47.4/47.2         \\
\textbf{Avg}                                                                           & 84.5/69.5          & 64.5/54.8        & 87.1/85.1          & 81.6/81.1         & 48.9/48.6         & 36.4/36.0          & 17.3/15.2            & 26.1/23.8          & 60.3/57.0            & 53.6/51.1       \\ \hline  
\end{tabular}}
\caption{The detailed experimental results for Llama3.}
\label{tab:re10}
\end{table*}

\renewcommand{\arraystretch}{2.5}
\definecolor{myblue}{rgb}{0.8, 0.85, 0.95}

\begin{table*}[]
\small
\setlength{\tabcolsep}{1.3mm}{
\begin{tabular}{ccccccccccc}
\hline
\multicolumn{11}{c}{\textit{\textbf{Gemma2}}}                                                                                                                                                                                                                                                                                                                                                                                                        \\ \hline
\multirow{2}{*}{\textbf{\begin{tabular}[c]{@{}c@{}}n-shot\\    (seed)\end{tabular}}} & \multicolumn{2}{c}{\textbf{Bragging}} & \multicolumn{2}{c}{\textbf{Complaint}} & \multicolumn{2}{c}{\textbf{Sarcasm}} & \multicolumn{2}{c}{\textbf{Rumour Stance}} & \multicolumn{2}{c}{\textbf{GossipCop}} \\ \cline{2-11} 
                                                                                     & \textbf{ICL}       & \textbf{IT}      & \textbf{ICL}       & \textbf{IT}       & \textbf{ICL}      & \textbf{IT}      & \textbf{ICL}         & \textbf{IT}        & \textbf{ICL}       & \textbf{IT}       \\ \hline
\textbf{1(42)}                                                                         & 83.1/69.1          & 83.6/68.7        & 86.1/85.5          & 85.9/85.4         & 46.4/46.3         & 59.7/57.9        & 52.1/38.6            & 57.1/41.1          & 63.0/59.3            & 72.7/63.6         \\
\textbf{1(43)}                                                                         & 71.2/59.4          & 82.3/67.9        & 83.5/83.0            & 87.4/86.8         & 43.1/43.1         & 56.6/55.3        & 35.2/28.2            & 55.8/39.3          & 64.9/59.5          & 72.7/63.6         \\
\textbf{1(44)}                                                                         & 74.2/60.7          & 86.9/71.5        & 81.9/81.6          & 85.7/85.2         & 39.6/39.6         & 54.1/53.2        & 54.1/38.8            & 54.2/41.4          & 57.8/55.2          & 72.7/63.6         \\
\textbf{1(45)}                                                                         & 80.0/66.3            & 83.6/68.6        & 86.2/85.7          & 87.7/87.1         & 35.0/34.6           & 58.4/56.9        & 43.6/36.5            & 54.8/40.9          & 55.9/54.7          & 74.4/56.6         \\
\textbf{1(46)}                                                                         & 78.5/65.8          & 87.8/73.1        & 82.3/82.0            & 80.9/80.6         & 42.9/42.9         & 57.3/55.9        & 42.5/35.2            & 56.6/40.8          & 43.8/43.8          & 73.7/63.7         \\
\textbf{Avg}                                                                           & 77.4/64.3          & 84.9/70.0          & 84.0/83.6            & 85.5/85.0           & 41.4/41.3         & 57.2/55.9        & 45.5/35.5            & 55.7/40.7          & 57.1/54.5          & 73.2/62.3         \\ \hline
\textbf{8(42)}                                                                         & 73.7/62.6          & 78.7/65.1        & 84.2/83.9          & 84.5/84.1         & 39.7/39.7         & 56.9/55.6        & 42.6/33.7            & 57.2/41.0            & 58.4/56.7          & 70.8/64.2         \\
\textbf{8(43)}                                                                         & 71.9/61.4          & 80.0/66.3          & 84.8/84.4          & 83.9/83.5         & 42.9/42.9         & 55.9/54.7        & 44.4/38.3            & 56.4/40.8          & 54.6/53.5          & 74.1/64.2         \\
\textbf{8(44)}                                                                         & 74.1/62.3          & 83.1/68.1        & 85.4/84.9          & 83.8/83.4         & 44.6/44.5         & 57.9/56.5        & 41.0/35.7              & 56.5/41.0            & 50.5/50.1          & 46.1/46.1         \\
\textbf{8(45)}                                                                         & 67.8/58.5          & 82.9/68          & 86.2/85.7          & 83.2/82.8         & 42.7/42.7         & 57.9/56.5        & 48.9/40.8            & 57.0/41.0              & 53.1/52.3          & 74.2/52.2         \\
\textbf{8(46)}                                                                         & 73.9/62.7          & 84.2/69.2        & 84.9/84.6          & 84.1/83.7         & 42.4/42.4         & 56.4/55.1        & 52.7/41.8            & 56.9/41.0            & 45.5/45.5          & 58.4/56.8         \\
\textbf{Avg}                                                                           & 72.3/61.5          & 81.8/67.4        & 85.1/84.7          & 83.9/83.5         & 42.5/42.5         & 57.0/55.7          & 45.9/38.1            & 56.8/40.9          & 52.4/51.6          & 64.7/56.7         \\ \hline
\textbf{16(42)}                                                                        & 66.8/57.6          & 82.8/68.3        & 87.7/87.1          & 86.4/85.9         & 50.7/50.3         & 58.2/56.7        & 44.5/34.1            & 56.9/40.9           & 51.7/51.2          & 72.2/62.2         \\
\textbf{16(43)}                                                                        & 70.6/60.3          & 81.3/67.5        & 85.5/85.1          & 84.8/84.4         & 47.0/46.9           & 56.6/55.3        & 45.4/38.9            & 56.9/40.8          & 55.0/53.8            & 45.7/45.7         \\
\textbf{16(44)}                                                                        & 68.2/58.4          & 82.9/68.4        & 84.9/84.6          & 82.3/82.0           & 49.9/49.6         & 56.5/55.3        & 45.8/36.0              & 55.3/40.0            & 58.8/57.0            & 55.0/53.8           \\
\textbf{16(45)}                                                                        & 63.5/55.0            & 83.1/68.3        & 86.8/86.3          & 82.0/81.7           & 47.5/47.4         & 54.7/53.7        & 45.0/37.5              & 56.9/40.7          & 55.4/54.4          & 72.9/42.2         \\
\textbf{16(46)}                                                                        & 73.6/62.0            & 83.3/68.5        & 88.0/87.5            & 83.3/83.0           & 45.7/45.6         & 56.4/55.2        & 50.6/40.4            & 57.0/40.8            & 52.0/51.5            & 56.7/55.3         \\
\textbf{Avg}                                                                           & 68.6/58.6          & 82.7/68.2        & 86.6/86.1          & 83.8/83.4         & 48.2/48.0           & 56.5/55.2        & 46.3/37.4            & 56.6/40.6          & 54.6/53.6          & 60.5/51.8         \\ \hline
\textbf{32(42)}                                                                        & 69.1/58.8          & 82.9/68.0          & 87.8/87.3          & 84.3/83.9         & 54.5/53.7         & 55.1/54.1        & 50.1/33.8            & 60.3/42.5           & 53.0/52.3            & 73.8/61.5         \\
\textbf{32(43)}                                                                        & 71.2/60.7          & 82.1/68.2        & 89.6/89.0            & 84.1/83.7         & 52.2/51.6         & 57.3/55.9        & 46.9/34.8            & 60.2/42.3          & 44.9/44.9          & 73.2/65.0           \\
\textbf{32(44)}                                                                        & 72.3/61.1          & 83.2/68.3        & 83.2/82.8          & 83.5/83.1         & 54.8/54.0           & 54.5/53.6        & 50.4/37.4            & 61.0/42.9            & 47.6/47.6          & 75.9/56.4         \\
\textbf{32(45)}                                                                        & 65.7/56.2          & 83.8/68.9        & 87.4/86.9          & 83.8/83.4         & 53.6/52.9         & 56.3/55.1        & 39.2/33.6            & 59.6/42.5          & 42.0/41.9            & 74.1/61.8         \\
\textbf{32(46)}                                                                        & 73.2/61.8          & 83.3/68.4        & 87.0/86.5            & 84.2/83.8         & 54.0/53.3           & 56.7/55.4        & 42.8/35.5            & 60.5/43.2          & 52.4/52.0            & 72.2/63.1         \\
\textbf{Avg}                                                                           & 70.3/59.7          & 83.1/68.4        & 87.0/86.5            & 84.0/83.6           & 53.8/53.1         & 56.0/54.8          & 45.9/35.0              & 60.3/42.7          & 48.0/47.7            & 73.8/61.5    \\ \hline    
\end{tabular}}
\caption{The detailed experimental results for Gemma2.}
\label{tab:re11}
\end{table*}

\renewcommand{\arraystretch}{2.5}
\definecolor{myblue}{rgb}{0.8, 0.85, 0.95}

\begin{table*}[]
\small
\setlength{\tabcolsep}{1.3mm}{
\begin{tabular}{ccccccccccc}
\hline
\multicolumn{11}{c}{\textit{\textbf{Phi-3}}}                                                                                                                                                                                                                                                       \\ \hline
\multirow{2}{*}{\textbf{\begin{tabular}[c]{@{}c@{}}n-shot\\    (seed)\end{tabular}}} & \multicolumn{2}{c}{\textbf{Bragging}} & \multicolumn{2}{c}{\textbf{Complaint}} & \multicolumn{2}{c}{\textbf{Sarcasm}} & \multicolumn{2}{c}{\textbf{Rumour Stance}} & \multicolumn{2}{c}{\textbf{GossipCop}} \\ \cline{2-11} 
                                                                                     & \textbf{ICL}       & \textbf{IT}      & \textbf{ICL}       & \textbf{IT}       & \textbf{ICL}      & \textbf{IT}      & \textbf{ICL}         & \textbf{IT}        & \textbf{ICL}       & \textbf{IT}       \\ \hline
\textbf{1(42)}                                                                         & 89.0/70.9            & 88.3/72.7        & 88.2/87.2          & 86.7/86.1         & 70.8/65.1         & 44.3/44.2        & 56.4/41.0              & 36.9/34.6          & 68.8/47.5          & 71.0/52.2           \\
\textbf{1(43)}                                                                         & 89.4/71.0            & 87.5/71.3        & 88.4/87.4          & 86.5/85.9         & 72.8/66.2         & 43.6/43.5        & 60.1/41.3            & 39.7/36.6          & 70.5/45.9          & 71.0/53.2           \\
\textbf{1(44)}                                                                         & 90.0/67.5            & 87.7/72.2        & 91.7/90.9          & 86.5/85.9         & 62.2/59.1         & 43.9/43.9        & 59.1/41.4            & 37.3/34.8          & 67.3/52.3          & 71.0/52.2           \\
\textbf{1(45)}                                                                         & 88.3/72.3          & 87.3/71.4        & 89.7/88.9          & 84.9/84.3         & 66.0/61.9           & 45.2/45.1        & 60.7/43.5            & 39.2/35.5          & 70.0/47.5            & 71.0/52.5           \\
\textbf{1(46)}                                                                         & 88.7/73.5          & 88.4/73.4        & 89.8/88.8          & 86.7/86.1         & 68.9/64.2         & 42.4/42.4        & 51.7/38.2            & 38.0/34.6            & 67.5/49.4          & 70.1/51.5         \\
\textbf{Avg}                                                                           & 89.1/71.0            & 87.8/72.2        & 89.6/88.6          & 86.3/85.7         & 68.2/63.3         & 43.9/43.8        & 57.6/41.1            & 38.2/35.2          & 68.8/48.5          & 70.8/52.3         \\ \hline
\textbf{8(42)}                                                                         & 88.7/73.5          & 88.1/72.4        & 90.1/89.4          & 86.5/86.0           & 59.1/57.1         & 44.2/44.1        & 46.4/34.4            & 36.1/33.3          & 69.6/50.0            & 71.4/53.0           \\
\textbf{8(43)}                                                                         & 88.5/74.0            & 88.3/73.8        & 89.1/88.4          & 86.5/85.9         & 63.1/60.0           & 43.6/43.6        & 50.0/37.9              & 36.0/34.0              & 69.0/51.3            & 70.5/52.2         \\
\textbf{8(44)}                                                                         & 89.6/73.3          & 87.7/72.6        & 89.9/89.0            & 86.5/85.9         & 68.7/64.1         & 44.7/44.6        & 51.9/38.3            & 38.6/36.0            & 68.7/52.0            & 70.6/52.1         \\
\textbf{8(45)}                                                                         & 89.8/73.8          & 88.1/73.0          & 90.0/89.3            & 86.5/86.0           & 56.4/54.7         & 44.3/44.2        & 53.3/39.9            & 37.4/34.8          & 69.2/52.7          & 71.4/53.0           \\
\textbf{8(46)}                                                                         & 87.8/74.1          & 87.3/71.6        & 89.3/88.5          & 85.5/84.9         & 65.8/61.8         & 44.0/44.0            & 57.5/41.0              & 37.9/35.1          & 71.9/51.6          & 71.4/53.0           \\
\textbf{Avg}                                                                           & 88.9/73.8          & 87.9/72.7        & 89.7/88.9          & 86.3/85.7         & 62.6/59.5         & 44.2/44.1        & 51.8/38.3            & 37.2/34.6          & 69.7/51.5          & 71.1/52.7         \\ \hline
\textbf{16(42)}                                                                        & 88.1/73.4          & 87.8/72.6        & 89.3/88.3          & 86.4/85.7         & 51.6/50.9         & 45.2/45.1        & 53.9/38.1            & 38.7/35.7          & 71.2/51.1          & 70.9/52.1         \\
\textbf{16(43)}                                                                        & 89.3/74.7          & 87.3/71.1        & 90.4/89.5          & 85.1/84.4         & 52.0/51.2           & 45.2/45.1        & 47.8/39.9            & 37.1/35.0            & 68.8/53.8          & 69.6/50.2         \\
\textbf{16(44)}                                                                        & 85.5/72.1          & 87.5/72.6        & 90.0/89.3            & 85.7/85.0           & 56.8/55.3         & 43.3/43.3        & 50.3/40.7            & 39.2/36.0            & 69.5/53.5          & 70.5/52.1         \\
\textbf{16(45)}                                                                        & 89.0/74.6            & 87.6/72.0          & 89.4/88.7          & 85.1/84.4         & 57.9/56.2         & 45.0/44.9          & 54.6/43.8            & 36.8/34.1          & 71.3/45.7          & 70.0/50.5           \\
\textbf{16(46)}                                                                        & 88.5/73.7          & 87.5/72.2        & 89.0/88.1            & 84.9/84.2         & 55.7/54.4         & 43.3/43.3        & 44.4/39.3            & 37.5/34.5          & 70.1/50.0            & 71.1/51.7         \\
\textbf{Avg}                                                                           & 88.1/73.7          & 87.6/72.1        & 89.6/88.8          & 85.4/84.8         & 54.8/53.6         & 44.4/44.3        & 50.2/40.4            & 37.9/35.1          & 70.2/50.8          & 70.4/51.3         \\ \hline
\textbf{32(42)}                                                                        & 86.1/72.4          & 87.8/72.4        & 89.0/88.1            & 84.0/83.4           & 48.9/48.5         & 43.6/43.6        & 46.0/36.9              & 37.3/34.6          & 71.1/49.2          & 70.8/51.5         \\
\textbf{32(43)}                                                                        & 87.8/73.4          & 87.3/71.3        & 88.6/87.9          & 85.1/84.4         & 48.2/47.8         & 43.8/43.8        & 60.9/34.2            & 36.6/34.1          & 70.4/52.6          & 69.9/50.1         \\
\textbf{32(44)}                                                                        & 87.2/73.1          & 87.8/73.0          & 89.3/88.6          & 84.8/84.2         & 51.5/50.9         & 44.1/44.1        & 51.5/38.2            & 35.7/34.2          & 72.2/49.3          & 69.9/49.7         \\
\textbf{32(45)}                                                                        & 87.2/73.2          & 87.5/72.2        & 89.0/87.9            & 85.4/84.7         & 47.3/47.1         & 44.8/44.7        & 53.2/37.3            & 38.6/36.0            & 71.6/46.3          & 70.8/50.9         \\
\textbf{32(46)}                                                                        & 83.3/69.4          & 88.1/72.7        & 89.1/88.4          & 85.8/85.2         & 50.8/50.3         & 43.4/43.4        & 43.9/37.7            & 36.3/33.8          & 70.6/51.5          & 69.6/50.3         \\
\textbf{Avg}                                                                           & 86.3/72.3          & 87.7/72.3        & 89.0/88.2            & 85.0/84.4           & 49.3/48.9         & 44.0/43.9          & 51.1/36.9            & 36.9/34.5          & 71.2/49.8          & 70.2/50.5  \\ \hline      
\end{tabular}}
\caption{The detailed experimental results for Phi-3.}
\label{tab:re12}
\end{table*}



\renewcommand{\arraystretch}{1.2}
\begin{table*}[bp]
\begin{tabular}{lccccccccccc}
\hline
\textbf{\multirow{2}{*}{Model}} & \textbf{\multirow{2}{*}{seed}} & \multicolumn{2}{c}{\textbf{Bragging}} & \multicolumn{2}{c}{\textbf{Complaint}} & \multicolumn{2}{c}{\textbf{Sarcasm}} & \multicolumn{2}{c}{\textbf{Rumour Stance}} & \multicolumn{2}{c}{\textbf{GossipCop}} \\
\cline{3-12} 
 &  & Acc & F1 & Acc & F1 & Acc & F1 & Acc & F1 & Acc & F1 \\
 \hline
\textbf{\multirow{6}{*}{Qwen2}} & 42 & 80.0 & 62.8 & 87.2 & 86.5 & 57.5 & 55.7 & 26.3 & 46.7 & 76.4 & 65.6 \\
 & 43 & 82.5 & 66.1 & 82.8 & 82.4 & 58.2 & 56.2 & 30.2 & 39.3 & 72.9 & 59.5 \\
 & 44 & 85.0 & 68.3 & 89.7 & 89.1 & 53.6 & 51.9 & 31.5 & 44.8 & 69.0 & 64.0 \\
 & 45 & 84.4 & 70.2 & 89.8 & 89.2 & 66.9 & 62.0 & 29.7 & 50.2 & 76.6 & 63.4 \\
 & 46 & 88.6 & 72.6 & 88.6 & 87.7 & 58.5 & 56.5 & 24.0 & 39.2 & 75.5 & 65.4 \\
 & Avg & 84.1 & 68.0 & 87.6 & 87.0 & 59.0 & 56.5 & 28.3 & 44.0 & 74.1 & 63.6 \\
  \hline
\textbf{\multirow{6}{*}{Baichuan2}} & 42 & 65.4 & 51.5 & 59.1 & 57.7 & 55.6 & 52.2 & 26.7 & 37.6 & 68.0 & 59.7 \\
 & 43 & 58.0 & 48.2 & 66.9 & 66.9 & 45.9 & 45.5 & 29.6 & 29.8 & 62.2 & 57.7 \\
 & 44 & 60.8 & 48.5 & 78.4 & 77.6 & 52.4 & 50.6 & 21.8 & 32.7 & 74.7 & 55.7 \\
 & 45 & 58.3 & 49.7 & 67.6 & 67.5 & 66.0 & 59.3 & 24.2 & 29.4 & 74.3 & 59.0 \\
 & 46 & 84.4 & 65.0 & 58.3 & 57.2 & 49.8 & 48.7 & 35.3 & 34.5 & 72.5 & 57.5 \\
 & Avg & 65.4 & 52.6 & 66.1 & 65.4 & 53.9 & 51.3 & 27.5 & 32.8 & 70.3 & 57.9 \\
 \hline
\textbf{\multirow{6}{*}{GLM4}} & 42 & 89.8 & 72.8 & 85.8 & 85.2 & 43.5 & 43.5 & 21.4 & 27.8 & 75.0 & 59.8 \\
 & 43 & 87.7 & 72.5 & 84.0 & 83.5 & 56.2 & 54.7 & 24.7 & 30.6 & 73.1 & 50.0 \\
 & 44 & 84.2 & 69.4 & 86.6 & 86.1 & 65.3 & 60.6 & 24.6 & 29.6 & 72.7 & 57.8 \\
 & 45 & 89.0 & 74.9 & 88.6 & 87.9 & 62.0 & 58.6 & 29.1 & 25.8 & 74.3 & 62.0 \\
 & 46 & 88.1 & 74.6 & 87.0 & 86.4 & 58.3 & 56.2 & 24.3 & 27.6 & 64.2 & 60.1 \\
 & Avg & 87.8 & 72.8 & 86.4 & 85.8 & 57.1 & 54.7 & 24.8 & 28.3 & 71.9 & 57.9 \\
 \hline
\textbf{\multirow{6}{*}{Llama3}} & 42 & 70.9 & 52.2 & 78.5 & 78.3 & 53.4 & 52.5 & 25.0 & 14.7 & 68.6 & 58.1 \\
 & 43 & 70.1 & 57.7 & 79.6 & 79.4 & 57.6 & 56.0 & 28.8 & 19.9 & 70.9 & 53.8 \\
 & 44 & 70.9 & 57.9 & 87.9 & 87.1 & 42.5 & 42.5 & 31.0 & 25.6 & 62.2 & 55.7 \\
 & 45 & 67.6 & 57.4 & 85.1 & 84.7 & 38.1 & 37.9 & 29.8 & 16.4 & 69.1 & 58.5 \\
 & 46 & 63.4 & 54.6 & 83.4 & 83.1 & 56.7 & 55.1 & 24.9 & 16.2 & 68.5 & 58.4 \\
 & Avg & 68.6 & 55.9 & 82.9 & 82.5 & 49.7 & 48.8 & 27.9 & 18.6 & 67.9 & 56.9 \\
 \hline
\textbf{\multirow{6}{*}{Gemma2}} & 42 & 67.5 & 55.8 & 84.1 & 83.6 & 39.2 & 39.2 & 31.8 & 38.6 & 63.5 & 60.9 \\
 & 43 & 55.8 & 46.3 & 84.3 & 83.8 & 45.6 & 45.5 & 25.1 & 28.2 & 70.5 & 62.1 \\
 & 44 & 56.5 & 47.6 & 84.1 & 83.6 & 39.5 & 39.5 & 27.8 & 38.8 & 67.1 & 62.2 \\
 & 45 & 56.5 & 49.7 & 85.1 & 84.6 & 32.8 & 31.9 & 25.0 & 36.5 & 72.7 & 63.7 \\
 & 46 & 32.6 & 31.6 & 82.9 & 82.5 & 33.1 & 32.4 & 26.4 & 35.2 & 56.5 & 55.3 \\
 & Avg & 53.8 & 46.2 & 84.1 & 83.6 & 38.0 & 37.7 & 27.2 & 35.5 & 66.1 & 60.8 \\
 \hline
\textbf{\multirow{6}{*}{Phi-3}} & 42 & 84.5 & 67.9 & 88.5 & 87.7 & 71.2 & 65.8 & 25.8 & 41.0 & 72.2 & 55.9 \\
 & 43 & 84.1 & 64.4 & 87.5 & 86.7 & 67.1 & 63.3 & 28.4 & 41.3 & 68.5 & 50.8 \\
 & 44 & 82.7 & 62.6 & 88.6 & 87.8 & 74.5 & 67.1 & 25.2 & 41.4 & 68.2 & 52.7 \\
 & 45 & 72.9 & 58.8 & 89.0 & 88.3 & 67.7 & 63.9 & 30.8 & 43.5 & 69.9 & 55.5 \\
 & 46 & 75.1 & 62.6 & 90.1 & 89.4 & 66.0 & 62.3 & 25.3 & 38.2 & 70.6 & 53.4 \\
 & Avg & 79.9 & 63.3 & 88.8 & 88.0 & 69.3 & 64.5 & 27.1 & 41.1 & 69.9 & 53.7
\\ \hline
\end{tabular}%
\caption{The detailed experimental results in the CoT strategy.}

\end{table*}


\begin{table*}[htbp]
\centering
\resizebox{\textwidth}{!}{%
\begin{tabular}{lcccccccccccc}
\hline
\multirow{2}{*}{\textbf{Model}} & \multicolumn{2}{c}{\textbf{Bragging}} & \multicolumn{2}{c}{\textbf{Complaint}} & \multicolumn{2}{c}{\textbf{Sarcasm}} & \multicolumn{2}{c}{\textbf{Rumour Stance}} & \multicolumn{2}{c}{\textbf{GossipCop}} & \multicolumn{2}{c}{\textbf{Avg}} \\
\cline{2-13} 
 & Acc & F1 & Acc & F1 & Acc & F1 & Acc & F1 & Acc & F1 & Acc & F1 \\
 \hline
\textbf{Qwen2} & 86.2 & 70.6 & 86.9 & 86.5 & 36.2 & 35.9 & 28.5 & 24.4 & 70.4 & 62.2 & 61.7 & 55.9 \\
\textbf{Baichuan2} & 35.9 & 34.6 & 64.7 & 58.2 & 76.3 & 46.1 & 45.8 & 30.4 & 52.1 & 50.8 & 55.0 & 44.0 \\
\textbf{GLM4} & 85.5 & 72.5 & 81.7 & 81.3 & 27.4 & 25.4 & 24.1 & 22.0 & 71.7 & 71.7 & 58.1 & 54.6 \\
\textbf{Llama3} & 73.4 & 60.3 & 89.3 & 88.3 & 43.5 & 43.5 & 14.9 & 16.3 & 72.8 & 54.1 & 58.8 & 52.5 \\
\textbf{Gemma2} & 60.9 & 52.2 & 83.5 & 83.1 & 48.3 & 48.0 & 53.6 & 39.3 & 75.3 & 63.0 & 64.3 & 57.1 \\
\textbf{Phi-3} & 90.8 & 71.7 & 86.6 & 86.0 & 35.3 & 34.9 & 37.7 & 35.0 & 72.0 & 45.8 & 64.5 & 54.7
\\ \hline
\end{tabular}%
}
\caption{The detailed experimental results in the zero-shot strategy.}
\end{table*}

\end{document}